\documentclass[journal]{IEEEtran}
\usepackage{graphicx}
\usepackage{subcaption}
\usepackage{multirow}
\usepackage[table]{xcolor}
\usepackage{color}
\usepackage{colortbl}
\usepackage{tabularx}
\usepackage{bm}
\usepackage{booktabs}
\usepackage{url}
\usepackage{stackengine}
\usepackage{float}
\usepackage{verbatim}
\usepackage{array}
\usepackage{cite}
\usepackage{algorithm}
\usepackage{algorithmic}
\usepackage[colorlinks,linkcolor=blue]{hyperref}
\usepackage{amsmath,amssymb} 
\usepackage{comment}

\begin{document}
\title{Adapting Segment Anything Model for Change Detection in VHR Remote Sensing Images}

\author{Lei~Ding, Kun Zhu, Daifeng Peng, Hao Tang, Kuiwu Yang and Lorenzo~Bruzzone,~\IEEEmembership{Fellow,~IEEE}~

\thanks{L. Ding, K. Zhu and K. Yang are with the Information Engineering University, ZhengZhou, China (E-mail: dinglei14@outlook.com, zkun@whu.edu.cn, yangkw@aliyun.com).}

\thanks{D. Peng is with the Nanjing University of Information Science and Technology, Nanjing, China (E-mail: daifeng@nuist.edu.cn.}

\thanks{H. Tang is with the Department of Information Technology and Electrical Engineering, ETH Zurich, 8092 Zurich, Switzerland. (E-mail: hao.tang@vision.ee.ethz.ch).}

\thanks{L. Bruzzone is with the Department of Information Engineering and Computer Science, University of Trento, 38123 Trento, Italy (lorenzo.bruzzone@unitn.it).}

\thanks{This document is funded by the National Natural Science Foundation of China (No. 42201443).}}

\markboth{Manuscript under review}%
{Shell \MakeLowercase{\textit{et al.}}: Bare Demo of IEEEtran.cls for IEEE Journals}

\maketitle

\begin{abstract}
Vision Foundation Models (VFMs) such as the Segment Anything Model (SAM) allow zero-shot or interactive segmentation of visual contents, thus they are quickly applied in a variety of visual scenes. However, their direct use in many Remote Sensing (RS) applications is often unsatisfactory due to the special imaging properties of RS images. In this work, we aim to utilize the strong visual recognition capabilities of VFMs to improve change detection (CD) in very high-resolution (VHR) remote sensing images (RSIs). We employ the visual encoder of FastSAM, a variant of the SAM, to extract visual representations in RS scenes. To adapt FastSAM to focus on some specific ground objects in RS scenes, we propose a convolutional adaptor to aggregate the task-oriented change information. Moreover, to utilize the semantic representations that are inherent to SAM features, we introduce a task-agnostic semantic learning branch to model the semantic latent in bi-temporal RSIs. The resulting method, SAM-CD, obtains superior accuracy compared to the SOTA fully-supervised CD methods and exhibits a sample-efficient learning ability that is comparable to semi-supervised CD methods. To the best of our knowledge, this is the first work that adapts VFMs to CD in VHR RS images.
\end{abstract}




\section{Introduction}\label{sc1}

Recently, vision foundation models (VFMs) have emerged and gained great research focus in the field of computer vision. Utilizing the knowledge gained in meta-scale datasets, foundation models such as Segment Anything Model (SAM)\cite{kirillov2023segment} and its variants (e.g., FastSAm\cite{zhao2023fastSAM}, Mobile SAM\cite{zhang2023mobilesam}) are able to recognize visual contents in a training-free manner and obtain fine-grained semantic masks. Their strong generalization across different imaging conditions and visual objects promotes greatly the applications in real-world scenarios. However, due to the inductive bias learned in natural images, foundational models exhibit limitations when applied to images in some specific domains, such as medical images and Remote Sensing Images (RSIs). According to the literature survey\cite{ji2023segment}, SAM pays more attention to the foreground objects and often fails to segment small and irregular objects. In this paper, we focus on adapting SAM to improve one of the fundamental tasks in RS, i.e., Change Detection (CD) of Very High-Resolution (VHR) RSIs.

CD is the task of segmenting content changes in multi-temporal RSIs. It is crucial for a wide range of real-world applications, including environmental monitoring, urban management, disaster alerting, land cover/land use (LCLU) monitoring, etc. With the continuous development of Earth Observation technologies and deep learning methods, we are now able to monitor Earth's surface with large volumes of HR RSIs and automatically detect the changes. In particular, deep neural networks, including Convolutional Neural Networks (CNNs) and Vision Transformers (ViTs) \cite{dosovitskiy2020image}, are widely used for CD in VHR RSIs. A common practice is to use weight-sharing CNNs, i.e., siamese CNNs, to aggregate and segment the multi-temporal semantic changes. ViTs are also leveraged as encoders to aggregate the features or to model the semantic-change correlations. State-of-the-art (SOTA) methods can detect changes with high accuracy.

Although much progress has been made, the use of CD techniques in real-world applications is still limited. The deep learning-based methods generally require large volumes of high-quality training data, whereas it is difficult to collect enough well-annotated change instances. The construction of a large CD training set requires that there is a certain time gap between the acquisition dates, and the observation platform covers large areas. For some small or rare LCLU classes, it is often difficult to collect enough training samples. Moreover, in multi-temporal VHR RSIs there are often differences in imaging conditions, such as illumination conditions, imaging angle, observation seasons, or even sensing platforms that are specific of each dataset. Therefore, given only change samples but no semantic information, it is still challenging to discriminate the semantic changes from temporal differences.

In this paper, we aim to utilize the strong and universal semantic exploitation capability of SAM to boost CD accuracy and reduce the dependence on large volumes of training samples. The major contributions can be summarized as follows:

\begin{enumerate}
    \item Introducing VFMs to the CD of VHR RSIs and proposing a Segment Anything Model-based Change Detection (SAM-CD) network. The SAM-CD adapts the FastSAM, an efficient variant of the SAM, to RS scenes and utilizes the pre-trained priors to embed multi-temporal LCLU representations. To the best of our knowledge, this is the first work that introduces and adapts VFMs to the CD task. Experimental results reveal that SAM-CD obtains significant accuracy improvements over the SOTA methods, and exhibits a sample-efficient learning ability that is comparable to semi-supervised CD methods.

    \item Introducing a task-agnostic learning of semantic latent in CD of VHR RSIs. Differently from conventional CD approaches that exploit only difference information, we propose to model CD in a multi-task learning perspective and introduce an explicit semantic learning branch in the SAM-CD. Furthermore, a temporal constraint-based learning objective is proposed to drive the network toward embedding consistent semantic representations in multi-temporal features. The resulting approach can better discriminate semantic changes and does not require explicit semantic supervision.
    
\end{enumerate}

The remainder of this paper is organized as follows. Sec.\ref{sc2} reviews the literature works related to CD and vision foundation models. In Sec.\ref{sc3} we introduce the proposed SAM-CD. Sec.\ref{sc4} reports the experimental settings and the results obtained on a benchmark dataset. Sec.\ref{sc5} draws conclusions to this study.
\section{Related Work}\label{sc2}
This section first reviews CD methods using deep neural networks and then introduces the development of VFMs.

\subsection{Change Detection in RSIs}

Before deep learning gained prominence, CD methods extracted and analyzed the change features to segment changes. They can be categorized into three categories based on the types of analyzed features, including texture features, object-based features, and angular features \cite{wen2021change}.

In recent years, CNNs have been extensively utilized for change detection due to their ability to capture contextual information in RSIs. Initially, deep CNN-based CD was treated as a segmentation task, where UNet-like CNN models are employed to directly segment the changed objects \cite{peng2019end}. In \cite{daudt2018fully}, a common CNN framework for CD is established. It utilizes two siamese CNN encoders to extract the multi-temporal features, before embedding them into change representations with a CNN decoder. 

The major challenges in CD on VHR RSIs are to distinguish semantic changes between seasonal changes and mitigating spatial misalignment as well as illumination differences. In CNN-based methods, channel-wise feature difference operations are commonly used to extract the change features \cite{daudt2018fully, zhang2020feature}. Another common strategy is to leverage multi-scale features to reduce the impact of redundant spatial details \cite{hou2021high}. Multi-scale binary supervisions are also introduced in \cite{peng2019end} to align the embedding of change features. Recently, the attention mechanism has been widely used to model the spatial context in RSIs \cite{cheng2023change}. Channel-wise attention is often used to improve the change representations \cite{li2022remote, peng2021scdnet}, while spatial attention is often used to exploit the long-range context dependencies \cite{chen2020dasnet, shi2021deeply}. Apart from these techniques, change augmentation (i.e. synthesizing change samples) is a commonly used strategy in single-temporal CD to supervise the learning of semantic changes. Typical approaches include generating change instances \cite{chen2023exchange} or removing objects \cite{seo2023self} in VHR RSIs.

Another research focus in CD is to model the temporal dependencies in the pairs of RSIs. In \cite{mou2018learning} hybrid CNN and Recurrent Neural Networks (RNN) are proposed for multi-class CD in RSIs, where the RNN modules are leveraged to reason the temporal information. In \cite{chen2019change} a multi-layer RNN module is adopted to learn change probabilities. In \cite{ding2022bi}, a cross-time self-attention is proposed to model the implicit semantic correlations in muti-temporal features. Graph convolutional networks are also an efficient technique to propagate LCLU information and thus are often adopted in semi-supervised \cite{wu2021multiscale} and unsupervised CD methods \cite{tang2021unsupervised}.

Recently Vision Transformers (ViTs) \cite{dosovitskiy2020image} gained great research interests in visual tasks. There are two strategies to utilize ViTs for CD of VHR RSIs. The first one is to replace CNN backbones with ViTs to extract temporal features, such as ChangeFormer\cite{yuan2022transformer} and EATDer\cite{ma2023eatder}. Meanwhile, ViTs can also be used to model the temporal dependencies. In BiT\cite{chen2021remote}, a transformer encoder is employed to extract changes of interest, while two siamese transformer decoders are placed to refine the change maps. In CTD-Former\cite{zhang2023relation}, a cross-temporal transformer is proposed to interact between the different temporal branches. In \cite{ding2022joint} a SCanFormer is introduced to model jointly the representation of semantic and change embeddings.

\subsection{Vision Foundation Models}

One of the bottleneck problems in applying deep neural networks in real-world applications is their requirement for a large amount of well-annotated training data, especially for dense prediction tasks such as semantic segmentation and CD. To address this problem, researchers have explored to pretrain vision models with web-scale datasets to obtain universal recognition capability that can be generalized to downstream tasks. A notable example is CLIP \cite{radford2021CLIP}, a model that can describe visual content with text descriptions, trained with 400 million image-text pairs. Its zero-shot image classification accuracy is comparable to a fully-supervised CNN. In \cite{bandara2022ddpm}, denoising diffusion probabilistic models are pre-trained on large-scale RS datasets before they serve as feature extractors in in CD of VHR RSIs.

An emerging trend in computer vision is to explore foundation models that can be generalized to common vision tasks with specified user prompts. This is sparked by the SAM \cite{kirillov2023segment}, a segmentation model that is trained on millions of annotated images to gain zero-shot generalization to 'unseen' images and objects. Given user prompts indicating the location or text descriptions, SAM is able to segment the objects with interests during the inference. Similar to SAM, SegGPT\cite{wang2023seggpt} also claims zero-shot recognition capability on common vision images. Furthermore, SEEM\cite{zou2023segment} enables more flexible user prompts including points, scribbles, text, audio, and their combinations. Since SAM is expensive in computation resources, FastSAM\cite{zhao2023fastSAM} is proposed to segment anything in real time. Its generalization performance is comparable to SAM, while its inference speed is 50 times faster than SAM.

Although VFMs claim to be able to 'segment anything' \cite{kirillov2023segment}, they exhibit significant limitations in certain domains, including medical images, manufacturing scenes, and RSIs ~\cite{ji2023segment}. Since VFMs are mostly trained on natural images, they tend to focus more on the foreground objects and struggle to segment small and irregular objects. Another limitation is that VFMs do not provide semantic classes associated with the masks. A feasible way to segment RSIs is to use text prompts to generate segmentation samples, and then progressively map all the classes \cite{osco2023segment}. In this work, we utilize adaption to fine-tune the VFM to learn semantic latent in VHR RSIs.

\section{Proposed Method}\label{sc3}

\begin{figure*}[t]
\centering
    \includegraphics[width=1\linewidth]{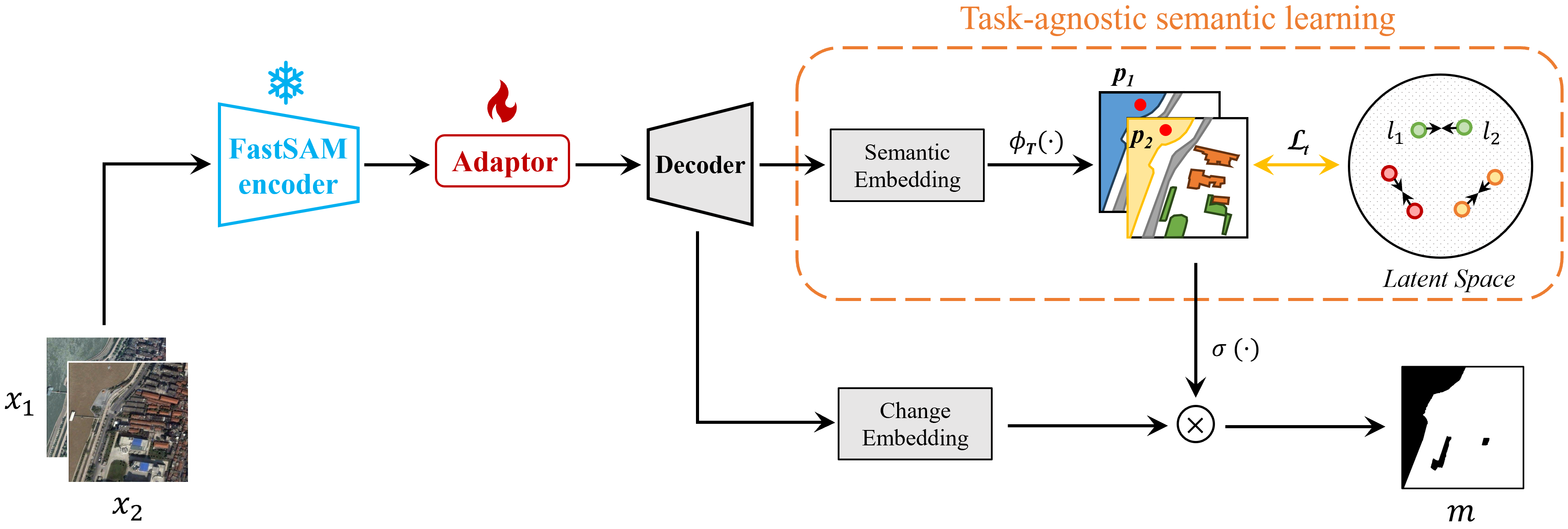}
    \caption{Architecture of the proposed SAM-CD.}
    \label{Fig.Overview}
\end{figure*}

This section describes the proposed SAM-CD architecture that adapts SAM to CD in VHR RSIs. First, we present an overview of the network. Then we elaborate on the technical details of the FastSAM adaptor and the task-agnostic semantic learning. Finally, we report the implementation details.

It is worth noting that the foundation model in SAM-CD is FastSAM since it supports access to low-level spatial features and requires fewer computational resources. However, it is possible to replace FastSAM with other vision foundation models in the SAM-CD architecture.

\subsection{Overview} \label{sc3.overview}

Typical deep learning-based methods for CD in RSIs leverage two siamese encoders to extract temporal features and then embed them into change representations with a shared decoder~\cite{daudt2018fully, peng2019end}. Let $(x_1, x_2)$ denote the pair of temporal images and $U(\cdot)$, $V(\cdot)$ denote the encoder and decoder networks. We can write:

\begin{equation}
    m = V[U(x_1, x_2)],
\end{equation}
where $m$ is a binary CD map that highlights the changes. A limitation of this architecture is that both $U(\cdot)$ and $V(\cdot)$ mainly focus on the temporal differences. Differently, we expect that the semantic changes can be learned better by comparing the underlying semantic features. With vision foundation models, it is now possible to extract the semantics of ground objects without categorical annotations. In SAM-CD, we extract the semantic latent to better discriminate the categorical changes. In other words, we employ VFM as visual encoders, denoted as $\widetilde{U}(\cdot)$, to extract universal semantic features (instead of temporal differences), while the $V(\cdot)$ compares the extracted features. This can be written as:
\begin{equation}
    m = V[\widetilde{U}(x_1), \widetilde{U}(x_2)],
\end{equation}

Fig. \ref{Fig.Overview} presents an overview of the proposed SAM-CD architecture. First, we leverage FastSAM as a frozen encoder to exploit the visual entities. For better generalization in RSIs, a trainable adaptor is introduced to adapt the extracted features. This is described in more detail in Sec.\ref{sc3.adaptor}. The obtained multi-scale FastSAM features are fused and upsampled in an unet-like convolutional decoder. Then, apart from the change branch that embeds change representations, we introduce an additional task-agnostic semantic learning branch to model the underlying semantic latent. This is illustrated in Sec.\ref{sc3.latent}. The resulting SAM-CD is semantic-aware, thus it can better capture the object changes in VHR RSIs.

\subsection{FastSAM Adaptor} \label{sc3.adaptor}

Although vision foundation models are capable of extracting semantic representations from any optical image, they exhibit limitations in discriminating ground objects in RSIs \cite{ji2023segment}. In the field of Natural Language Processing \cite{wu2023medical}, adaption is a commonly used strategy to fine-tune the pre-trained models to downstream tasks. Since the FastSAM is constructed with CNNs, we adopt convolutional adaptors to adapt the extracted features.

First, we aggregate the features extracted by FastSAM at the spatial scales of $1/32, 1/16, 1/8$ and $1/4$, denoted as ${f_1, f_2, f_3, f_4}$. Each feature $f_i$ is processed by a corresponding adaptor $\alpha$, denoted as:

\begin{equation}
    f_i^* =\alpha(f_i)=\gamma\{bn[conv(f_i)]\},
\end{equation}
where $conv$ denotes a 1$\times$1 convolutional layer, $bn$ denotes a batch normalization function, and $\gamma()$ is a RELU function. Since there are fewer object categories in RSIs than those in natural images, we reduce the channel number of $f_i^*$ to reduce redundancy.

In CD of VHR RSIs, low-level features are important for segmenting the areas with changes \cite{peng2019end, daudt2018fully}. Therefore, we employ an unet-like decoder to fuse the multi-scale features after adaptation. For each level of feature $f_i$, we fuse it with the lower level feature $f_{i+1}$ in a decoder block, denoted as:

\begin{equation}
    d_1 = f_1,
    d_{i+1} = conv[f_{i+1}, upsample(d_i)],
\end{equation}
where $d_i$ is the $i$-th layer of feature in the decoder. We then concatenate the resulting features $\{d_1, d_2, d_3\}$ to obtain semantic representations that are adaptive to the RS domain.

\subsection{Task-agnostic Semantic Learning}\label{sc3.latent}

Literature reveals that joint learning of multiple related tasks can improve the performance of each single task ~\cite{ruder2017overview, ding2020diresnet}. To boost the performance of CD, we introduce an additional temporal semantic learning branch. This is similar to the strategy of post-classification CD used in the last decades \cite{wu2017post}, i.e., segmenting changes after classifying the multi-temporal RSIs. However, in the proposed SAM-CD architecture, the learning of semantics and changes are conducted simultaneously.

In Sec.\ref{sc3.adaptor} we described the adapted SAM features $\{d_1, d_2, d_3\}$. We further use convolutional operations to transform them into a candidate latent $\hat{l}\in\mathbb{R}^{k\times w\times h}$ where $k$ indicates the number of interesting semantic clusters in RSIs. Supervising the learning of this latent with change labels will drive the network to focus on only the change classes, i.e., being task-specific. Instead, we use the underlying temporal constraints to supervise the learning of semantic latent.

Differently from literature research that explicitly supervises the learning of LCLU categories \cite{ding2022bi}, in binary CD task the semantic labels of each acquisition date are not available. Therefore, the SAM-CD implicitly supervises the learning of bi-temporal latent by aligning their feature representations. For each candidate latent $\hat{l}$, a softmax function $\phi$ is used to normalize them:

\begin{equation}\label{form.softmax}
    \phi_T(\hat{l}_i) = \frac{e^{\hat{l}_i/T}}{\sum_{j=1}^{n}e^{\hat{l}_j/T}},
\end{equation}
where $T$ is the temperature parameter to control the probability distribution of output features. We set $T>1$ to obtain more diverse semantic representations. The selection of the value of $T$ is discussed in Sec.\ref{sc4}.
Let $\{l_1, l_2\}$ be the normalized bi-temporal latent, we expect their semantic representations to be similar in the unchanged areas. Therefore, we propose a temporal constraint loss $\mathcal{L}_t$ to measure their temporal similarity, calculated as:

\begin{equation}\label{form.lt}
    \mathcal{L}_t(l_1, l_2) = [1-cosine(l_1, l_2)] \cdot c,
\end{equation}
where $c$ is the ground truth (GT) change label on which unchanged areas are annotated as 1. This is to exclude the changed areas from loss calculations. In this way, the task-agnostic semantic representations relevant to the RS domain can be exploited. We further utilize attention operations to embed the semantic focus into change features $f_c$, before mapping them into a change map $m$. The change features are obtained by forwarding the SAM features ${d_1, d_2, d_3}$ into a convolutional block. The attention embedding operation is as follows:

\begin{equation}\label{form.lt}
    m = conv_2\{\sigma [conv_1 (l_1 \oplus l_2)] \cdot f_c\}
\end{equation}
where $\oplus$ is a channel-wise concatenation operation, $\sigma$ is a sigmoid normalization function, $conv_1$ and $conv_2$ are two convolutional modules to adjust the feature channels. This ensures the CD results are semantic-aware, thus better segmenting the semantic changes.

\begin{figure}[t]
\centering
    \includegraphics[width=0.8\linewidth]{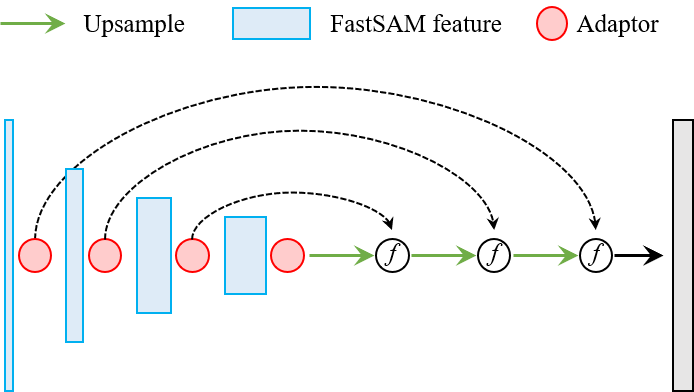}
    \caption{The proposed adaptor network to utilize FastSAM features. Each $\textcircled{\textit{f}}$ denotes a convolutional fusion operation.}\label{Fig.sem_extract}
\end{figure}

\subsection{Implementation Details}\label{sc3.detail}

In the following, we report the network settings and the training details in the implementation of the SAM-CD.

1) \textbf{SAM-CD}. The FastSAM is constructed using the yolo-v8 architecture. To obtain higher accuracy, we utilize the default version of FastSAM. Its parameter size is 68 Mb. Since the CD task is quite different from typical segmentation tasks, SAM-CD utilizes only the visual encoder of FastSAM and discards the prompt decoders. In the task-agnostic semantic learning branch, the semantic embedding block is a single 1$\times$1 convolutional layer. The number of semantic channels $k$ is empirically set to 8, considering that there are usually few interesting LCLU classes in typical CD applications. In the change detection branch, we leverage 6 layers of residual convolutional blocks to construct the change embedding module following the practice in \cite{ding2022bi}. This is to better transform semantic features into change representations.

3) \textbf{Training Settings}. The proposed method is implemented in PyTorch. The training is iterated for 50 epochs. We set the initial learning rate to 0.1 and update it at each iteration as $0.1*(1-iterations/total\_iterations)^{1.5}$. The Stochastic Gradient descent algorithm is used to optimize the gradient. We apply only simple geometric augmentations to the input images, including random flipping and random cropping. During the inference, we apply a test-time augmentation operation which includes 8 times of flipping operations to produce more stable prediction results.

For more details, readers are encouraged to visit our codes released at: \url{https://github.com/ggsDing/SAM-CD}.
\section{Experimental Results}\label{sc4}

This section presents the experimental results obtained on the benchmark datasets. First, we conduct ablation studies to test the effectiveness of the proposed SAM-CD architecture. Then, we test the label-efficient learning capability of the proposed method. Finally, we compare our results with the SOTA methods in CD on VHR RSIs.

\subsection{Experimental Setup}

\subsubsection{Dataset}

We conduct experiments on 4 benchmark CD datasets, including the LEVIR-CD dataset\cite{Chen2020}, WHU-CD dataset\cite{ji2018fully}, CLCD dataset, and S2Looking dataset\cite{shen2021s2looking}. Below we briefly introduce these two datasets.

\textbf{LEVIR-CD Dataset}:
This is a large-scale benchmark dataset for CD in VHR RSIs. It consists of Google Earth images collected in 6 cities in Texas, US between 2012 and 2016. Most of the changes in this dataset are related to construction growth. A total of 31,333 changed objects are annotated. The spatial resolution of images is 0.5m per pixel. There are a total of 637 pairs of images, each has 1024$\times$1024 pixels. The number of image pairs in the training, validation and test sets are 445, 64 and 128, respectively.

\textbf{WHU-CD Dataset}:
This is an aerial benchmark dataset constructed for building change detection. The RSIs were collected in Christchurch, New Zealand between 2012 and 2016, covering an urban area of 20.5 $km^2$. The observed area went through an earthquake in 2011, thus there is significant growth in the number of constructed buildings (from 12,796 to 16,077). The spatial resolution of images is 0.2m per pixel. The original image tiles have 32507$\times$15354 pixels. Following \cite{zhang2023relation}, we divide the dataset into training, validation and test sets with 6096, 762 and 762 pairs of image patches (each with a spatial size of 256$\times$256, respectively.

\textbf{CLCD Dataset}:
This dataset is constructed focusing on cropland changes. It consists of 600 pairs of Gaofen-2 satellite images collected in Guangdong Province, China. The two acquisition dates are in 2017 and 2019. Each image has 512$\times$512 pixels and the spatial resolution is 0.5 to 2m. Multiple types of LCLU changes are annotated, including buildings, roads, lakes and bare lands. The split of training, validation and test sets is 320, 120 and 120 image pairs, respectively.

\textbf{S2Looking Dataset}:
This is a large-scale benchmark dataset for building CD. It is constructed using VHRs collected by various satellites including GaoFen, SuperView and BeiJing-2, covering various rural areas around the world. There are a total of 5000 pairs of images, each has 1024$\times$1024 pixels. The spatial resolution of images is 0.5 to 0.8m. Different from other CD datasets, many image pairs in the S2Looking dataset are captured at different imaging angles. This causes spatial misalignment and thus poses challenges to CD.

\subsubsection{Evaluation Metrics}

Following literature studies\cite{zhang2023relation}, we adopt precision ($Pre$), Recall ($Rec$), $F_1$, intersection over union ($IoU$), and overall accuracy ($OA$) as the evaluation metrics to assess the accuracy of CD methods. The related calculations are:

\begin{equation}
    Pre=\frac{TP}{TP+FP},\quad Rec=\frac{TP}{TP+FN},
\label{formular_PR}
\end{equation}
\begin{equation}
    F_1=2\times\frac{P \times R}{P+R},\quad OA=\frac{TP+TN}{TP+FP+TN+FN},
\label{formular_F1OA}
\end{equation}
\begin{equation}
    IoU=\frac{TP}{TP+FP+FN},
\label{formular_IoU}
\end{equation}
where $TP$, $FP$, $TN$, and $FN$ represent true positive, false positive, true negative, and false negative, respectively.
It is worth noting that the quantitative results in this section are obtained under the same evaluation settings (including the implementation of calculations) as those in \cite{zhang2023relation}.

\begin{table*}[htbp]
\centering
    \caption{Quantitative results of the ablation study (tested on the Levir-CD dataset).}
    \resizebox{0.8\linewidth}{!}{%
        \begin{tabular}{l|cc|ccc}
        \toprule
            \multirow{2}*{Methods} & \multicolumn{2}{c|}{Proposed Techniques}  & \multicolumn{3}{c}{Accuracy}\\
            \cline{2-6}
            & SAM & $\mathcal{L}_{t}$ & $OA$(\%) & $mIoU$(\%) & $F_1$(\%) \\
            \hline
            siamese CNN (baseline) &  &  & 98.82 & 88.96 & 93.87 \\
            \hline
            SAM-CD (w/o. $\mathcal{L}_{t}$) &  $\surd$ & & 99.05 & 90.79 & 94.98 \\
            SAM-CD & $\surd$ & $\surd$ & 99.14 & 91.68 & 95.50 \\
        \bottomrule
        \end{tabular} \label{Table.Ablation} }
\end{table*}

\subsection{Ablation Study}

To quantitatively evaluate the improvements brought by the proposed techniques, we compare the obtained accuracy with the baseline method. It is a network constructed following the common settings in CD, i.e., using two siamese CNNs to extract temporal features and a decoder to embed the change representations as in \cite{daudt2018fully}. 
We construct the baseline CNN with two siamese ResNet34 encoders since their parameter size (81 Mb) is close to the FastSAM (68 Mb).

\begin{figure*}[!htb]
\centering
    {\includegraphics[height=0.5cm]{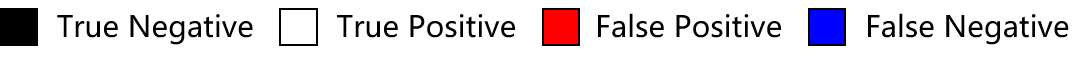}}\\
    \setlength{\tabcolsep}{1pt}
    \begin{tabular}{>{\centering\arraybackslash}m{0.4cm}>{\centering\arraybackslash}m{3.0cm}>{\centering\arraybackslash}m{3.0cm}>{\centering\arraybackslash}m{3.0cm}>{\centering\arraybackslash}m{3.0cm}>{\centering\arraybackslash}m{3.0cm}>{\centering\arraybackslash}m{3.0cm}}
        (a)&
        \includegraphics[width=3.0cm]{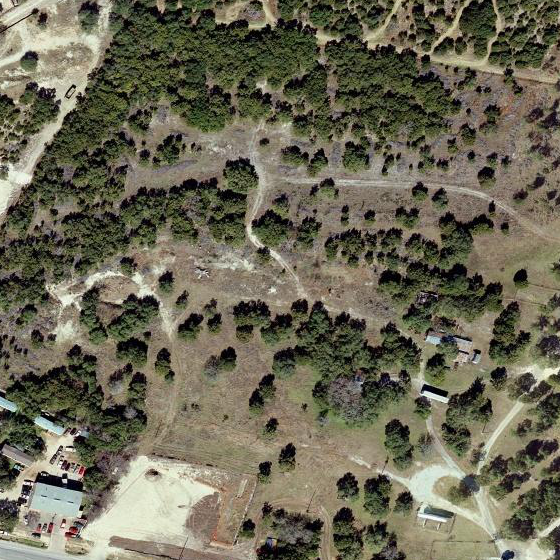} &
        \includegraphics[width=3.0cm]{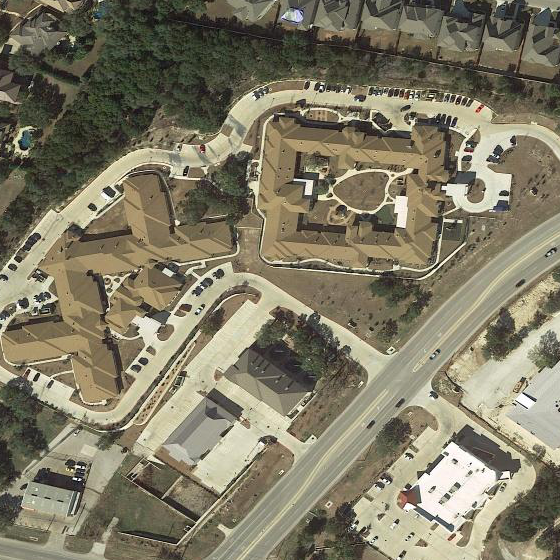} &
        \includegraphics[width=3.0cm]{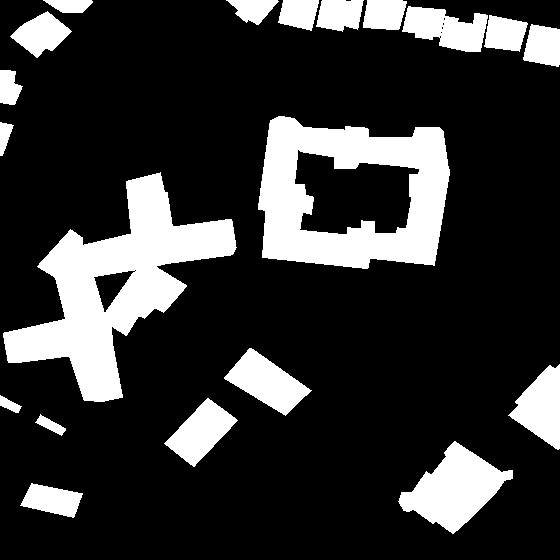} &
        \includegraphics[width=3.0cm]{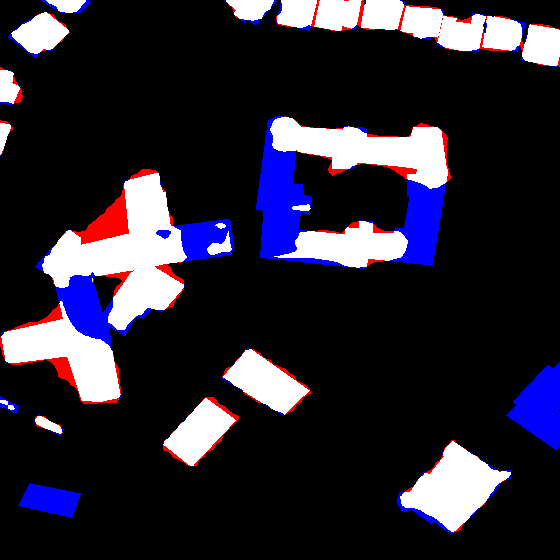} &
        \includegraphics[width=3.0cm]{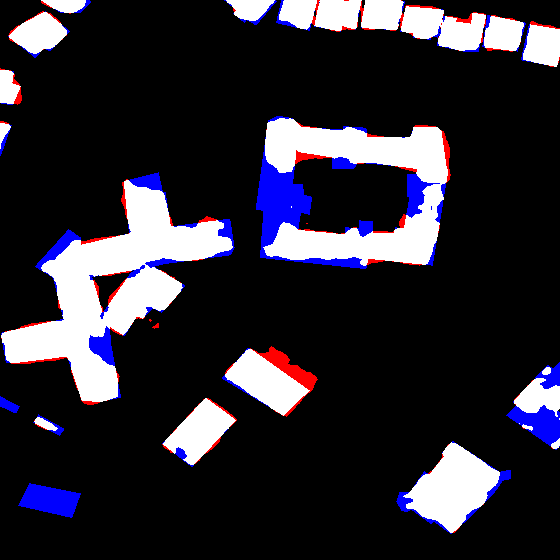} &
        \includegraphics[width=3.0cm]{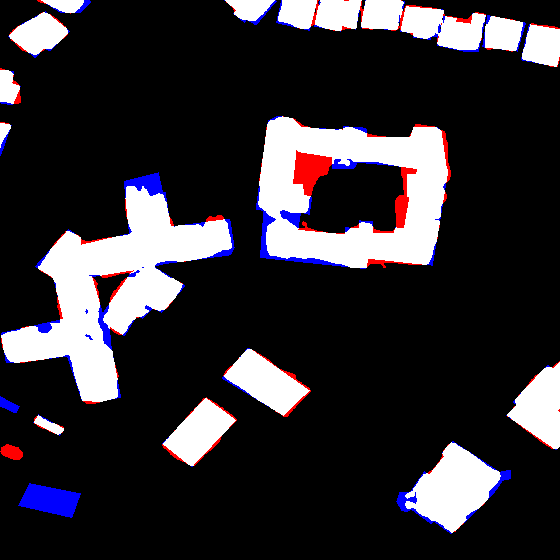}\\
        (b)&
        \includegraphics[width=3.0cm]{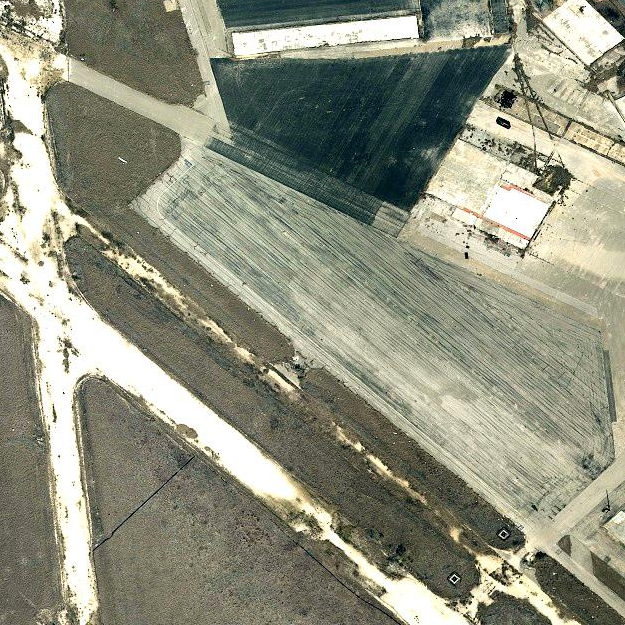} &
        \includegraphics[width=3.0cm]{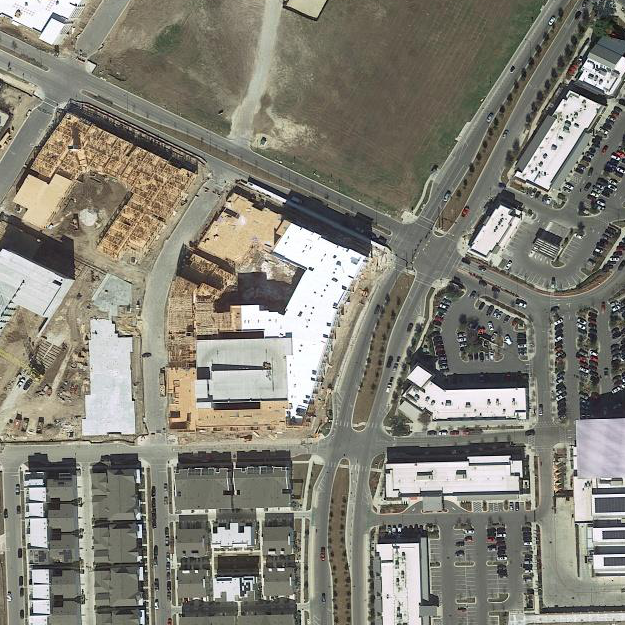} &
        \includegraphics[width=3.0cm]{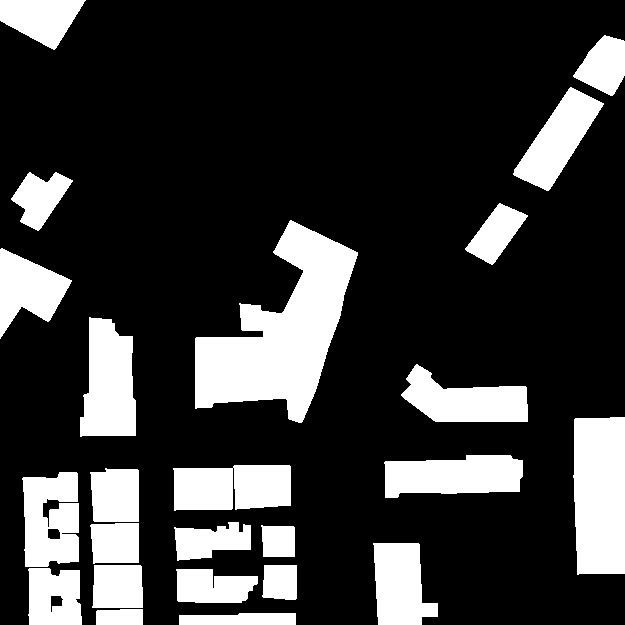} &
        \includegraphics[width=3.0cm]{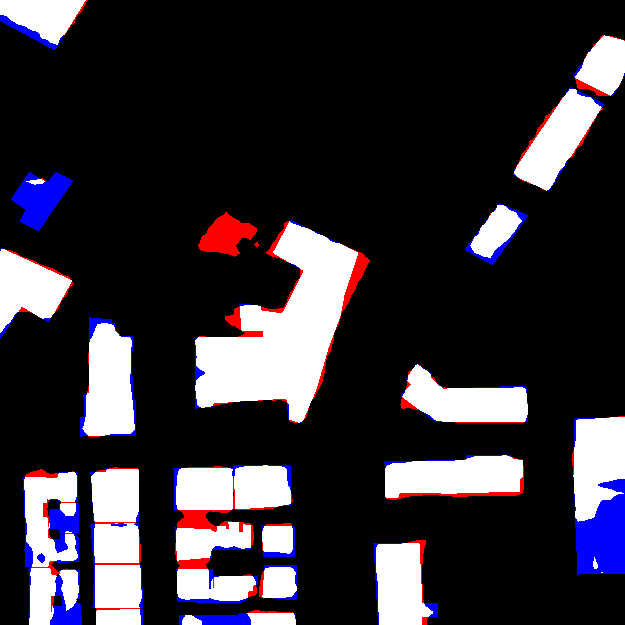} &
        \includegraphics[width=3.0cm]{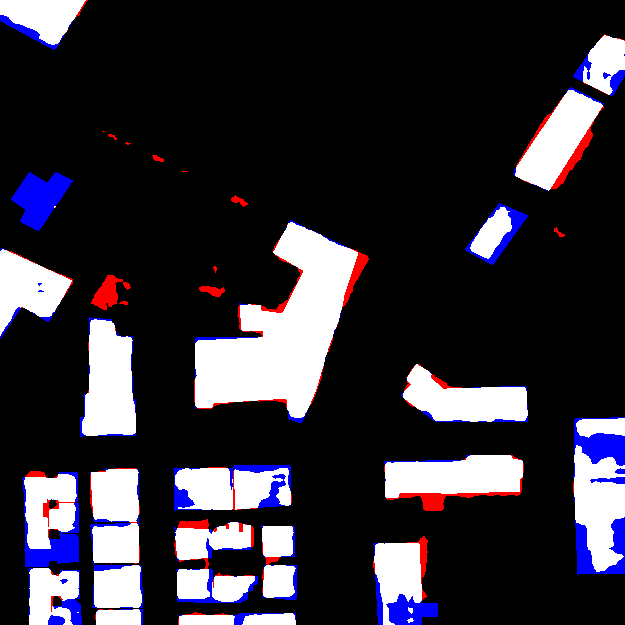} &
        \includegraphics[width=3.0cm]{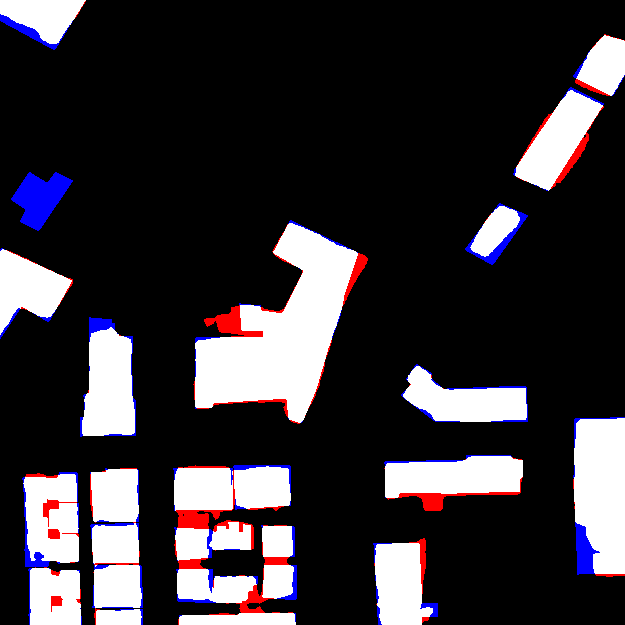}\\
        (c)&
        \includegraphics[width=3.0cm]{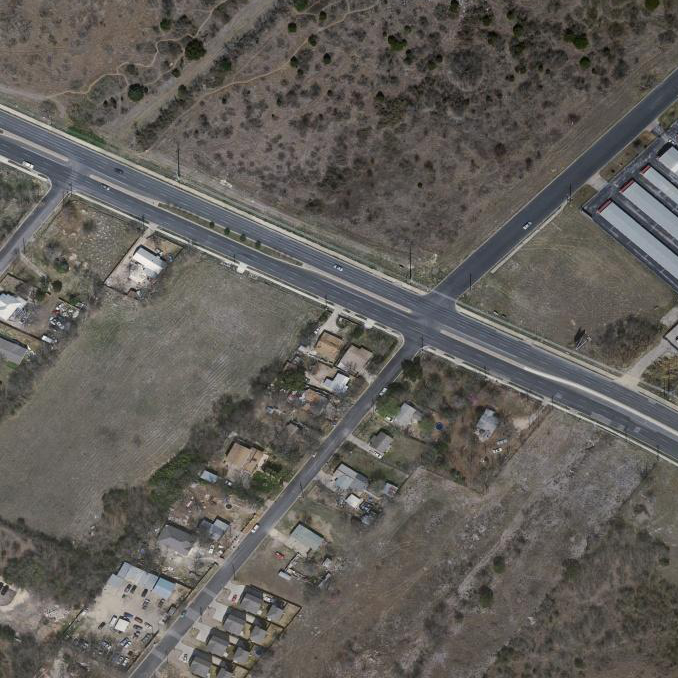} &
        \includegraphics[width=3.0cm]{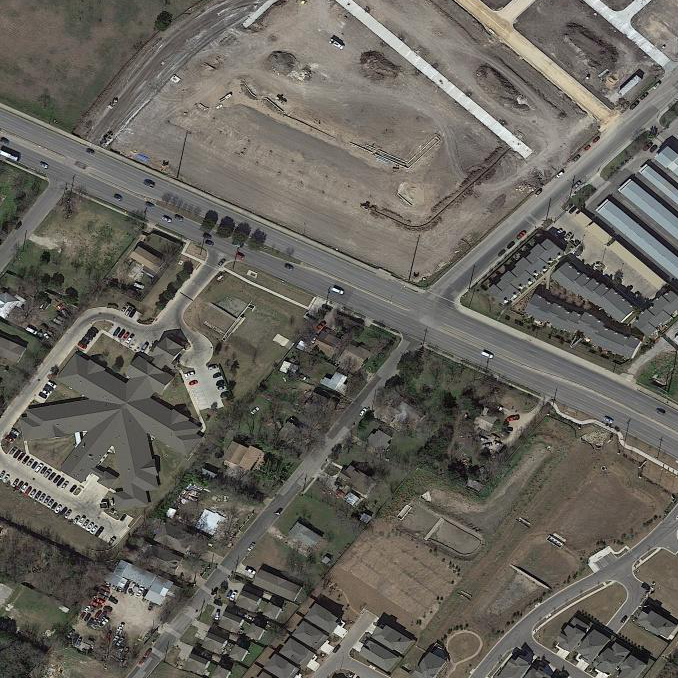} &
        \includegraphics[width=3.0cm]{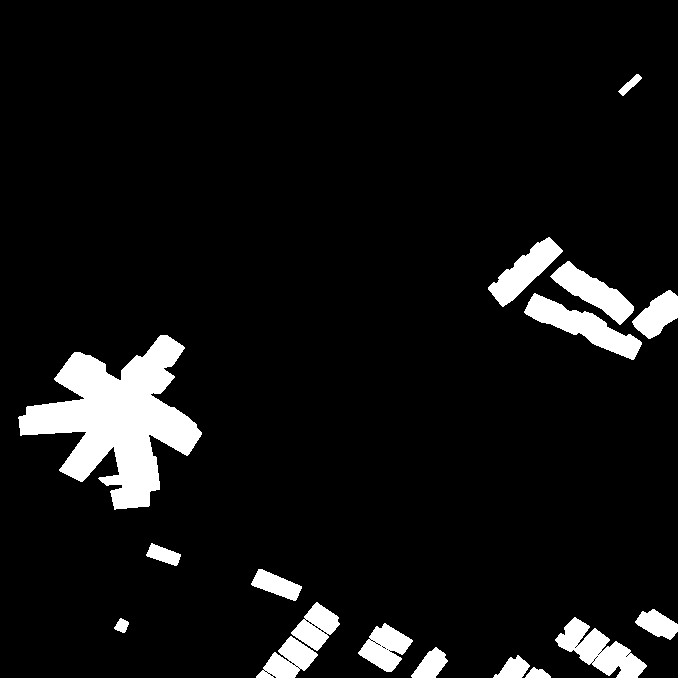} &
        \includegraphics[width=3.0cm]{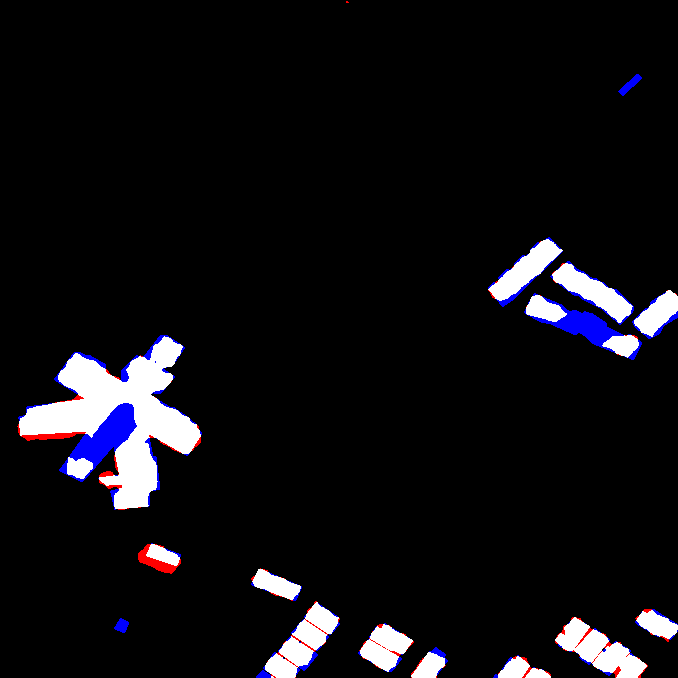} &
        \includegraphics[width=3.0cm]{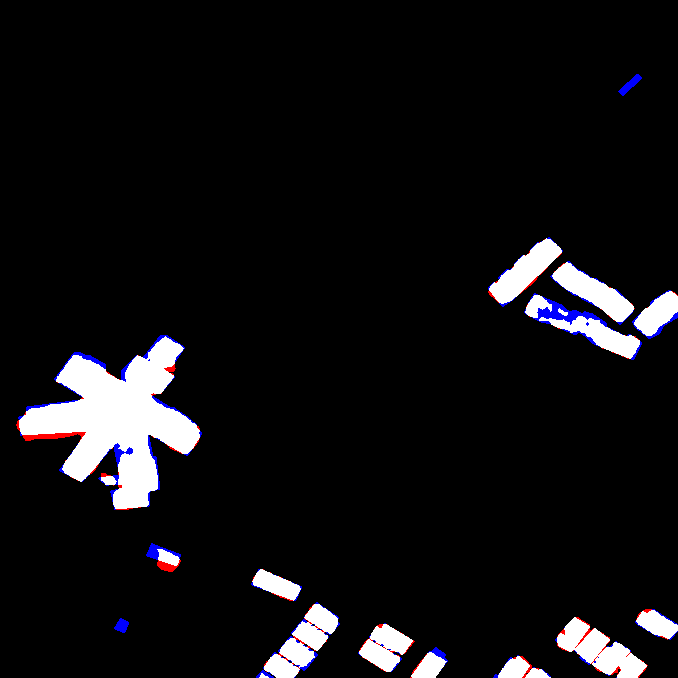} &
        \includegraphics[width=3.0cm]{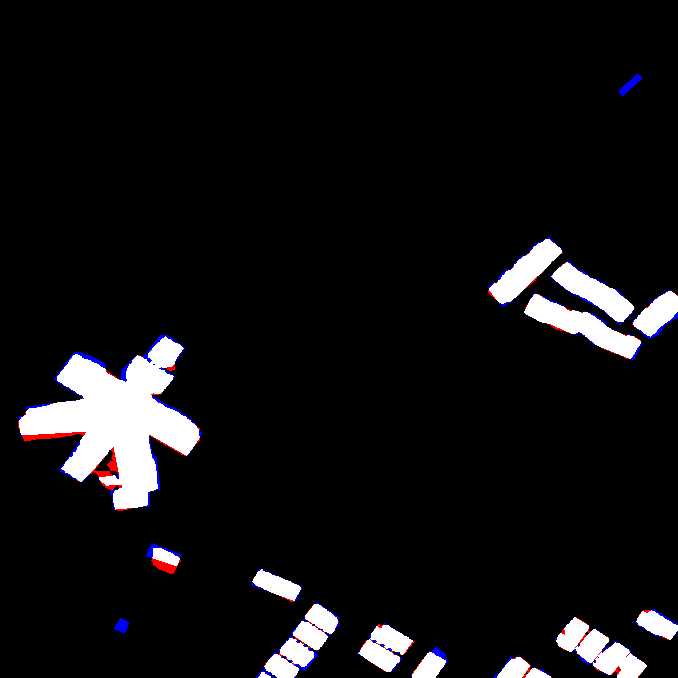}\\
        & $T_1$ image & $T_2$ image & GT & baseline & proposed SAM-CD (w/o. $\mathcal{L}_{t}$) & proposed SAM-CD\\
    \end{tabular}
    \caption{CD results of the different methods in the ablation study. The predicted maps are compared with the GT maps. The differences are highlighted in color.}
    \label{Fig.AblationResults}
\end{figure*}

The results are presented in Table \ref{Table.Ablation}. One can observe that compared to the baseline method, adding a plain FastSAM brings significant accuracy improvements. The $OA$, $mIoU$ and $F_1$ increase by $0.23\%$, $1.83\%$ and $1.11\%$, respectively. We also present some examples of the CD results in Fig.\ref{Fig.AblationResults}. One can observe that the proposed techniques in the SAM-CD bring progressive improvements to the segmentation results. Specifically, they boost the recognition of non-salient changes, e.g., the dark buildings in Fig.\ref{Fig.AblationResults}(a)(c), and the emerged objects that can be confused with the environment, e.g., the constructions in Fig.\ref{Fig.AblationResults}(b).

\textbf{SAM and FastSAM:} In the SAM-CD architecture, the FastSAM encoder can be replaced by other VFMs. Since SAM is the first and the most well-known VFM in the segmentation task, we also tested incorporating SAM encoder in the SAM-CD architecture. Table \ref{Table.sam_vs_fastsam} reports the performance of the SAM-CD equipped with different versions of the SAM and the FastSAM. Since the SAM is built on ViT and requires a fixed spatial input size of 1024$\times$1024 (smaller images will be upsampled before processing), it is much more expensive in computational resources compared to the FastSAM (built on CNNs). The costs of computer resources are measured by the number of parameters (param) and the Floating Point of OPerations (FLOPs). The FLOPs is tested using a pair of input images with a size of 3$\times$256$\times$2. The computational costs of the SAM-CD with SAM encoders are more than 100 times higher than those of the SAM-CD with FastSAM encoders in FLOPs.

Due to the limitation of computational resources, we only tested the accuracy of the SAM-b and the SAM-l encoders. One can observe that the SAM-CD (with FastSAM encoders) also exhibits better accuracy compared to their counterparts using the SAM encoders. Although the SAM is known to have better capability to exploit image semantics, the FastSAM can provide access to its low-level spatial features, which is crucial for the CD task. Since the FastSAM-x (default encoder in the FastSAM) leads to the highest accuracy, we adopt it as the default visual encoder in the SAM-CD. In pursuit of higher accuracy, readers are also encouraged to apply other VFMs as $\widetilde{U}(\cdot)$ in the SAM-CD architecture.

\begin{table*}[t]
    \centering
    \caption{Performance of the SAM-CD using different visual encoders (tested on the Levir-CD dataset).}
    \resizebox{0.8\linewidth}{!}{%
        \begin{tabular}{r|cc|ccc}
        \toprule
            \multirow{2}*{Method} & \multicolumn{2}{c|}{Computation Costs} & \multicolumn{3}{c}{Accuracy} \\
            \cline{2-6}
            & Param(Mb) & FLOPs(Gbps) & $OA$ & $F_{1}(\%)$ & $mIoU(\%)$ \\
            \hline
            SAM-CD (SAM-b) & 95.32 & 807.25 & 96.80 & 62.29 & 53.74 \\
            SAM-CD (SAM-l) & 330.97 & 2854.39 & 98.76 & 77.31 & 68.24 \\
            SAM-CD (SAM-h) & 685.90 & 5941.31 & -- & -- & -- \\
            SAM-CD (FastSAM-s) & 13.67 & 8.50 & 98.97 & 94.76 & 90.42 \\
            SAM-CD (FastSAM-x)  & 70.59 & 8.60 & \textbf{99.14} & \textbf{95.50} & \textbf{91.68} \\
        \bottomrule
        \end{tabular}
        }\label{Table.sam_vs_fastsam}
\end{table*}

\textbf{Selection of $T$:} In equation (\ref{form.softmax}), $T$ in the softmax function is an important parameter to control the diversity of semantic representations. We conduct experiments to select the optimal $T$ and report the results in Table \ref{Table.T_select}. One can observe that the increase of $T$ results in variations in the accuracy metrics. The optimal accuracy is obtained with $T=3$. This is in line with our intuition that a relatively large $T$ results in more softened latent distributions and thus better presents the semantics. Therefore, we set $T=3$ as the default value in the SAM-CD.

\begin{table}[t]
    \centering
    \caption{The accuracy of SAM-CD obtained with different $T$ values (tested on the Levir-CD dataset).}
    \resizebox{0.8\linewidth}{!}{%
        \begin{tabular}{r|ccc}
        \toprule
            SAM-CD & \multicolumn{3}{c}{Accuracy} \\
            \cline{2-4}
            $T$ values & $OA(\%)$ & $F_{1}(\%)$ & $mIoU(\%)$ \\
            \hline
            $T=1$ & 99.10 & 95.28 & 91.30 \\
            $T=2$ & 99.11 & 95.38 & 91.47 \\
            $T=3$ & \textbf{99.14} & \textbf{95.50} & \textbf{91.68} \\
            $T=4$ & 99.09 & 95.26 & 91.26 \\
            $T=5$ & 99.12 & 95.37 & 91.45 \\
        \bottomrule
        \end{tabular}
        }\label{Table.T_select}
\end{table}

\textbf{Selection of feature layers:} In the SAM-CD architecture, the FastSAM encoder extracts 4 layers of features corresponding to 1/4, 1/8, 1/16 and 1/32 of the input spatial size. Denote these features as $l_1, l_2, l_3$ and $l_4$, respectively. We further experiment to quantitatively evaluate the contribution of each layer of feature and to decide the optimal group of features to use in CD. This is conducted by slightly changing the SAM-CD architecture to use only one layer or a group of layers. The results are reported in Table \ref{Table.layer_select}. One can observe that performing CD using $l_1$ and $l_4$ leads to a relatively high accuracy. This reveals that both semantic features and spatial features are crucial to CD. Additionally, as indicated in the results using $l_2, l_3, l_4$, missing the low-level features ($l_1$) leads to a significant decrease in accuracy. Since the fusion of all 4 layers of features leads to the highest accuracy, this combination is selected in the SAM-CD architecture.

\begin{table}[t]
    \centering
    \caption{The accuracy of SAM-CD obtained using different layers of the FastSAM features (tested on Levir-CD).}
    \resizebox{0.8\linewidth}{!}{%
        \begin{tabular}{r|ccc}
        \toprule
            SAM-CD & \multicolumn{3}{c}{Accuracy} \\
            \cline{2-4}
            feature layers & $OA(\%)$ & $F_{1}(\%)$ & $mIoU(\%)$ \\
            \hline
            $l_1 (scale=1/4)$ & 99.02 & 94.82 & 90.53 \\
            $l_2 (scale=1/8)$ & 97.10 & 84.17 & 75.34 \\
            $l_3 (scale=1/16)$ & 98.34 & 91.15 & 84.73 \\
            $l_4 (scale=1/32)$ & 99.02 & 94.83 & 90.54 \\
            \hline
            $l_1, l_2, l_3$ & 99.12 & 95.44 & 91.57 \\
            $l_2, l_3, l_4$ & 98.84 & 93.92 & 89.05 \\
            $l_1, l_2, l_3, l_4$ & \textbf{99.14} & \textbf{95.50} & \textbf{91.68} \\
        \bottomrule
        \end{tabular}
        }\label{Table.layer_select}
\end{table}

\begin{figure}[!htb]
\centering
    \setlength{\tabcolsep}{1pt}
    {\includegraphics[height=0.5cm]{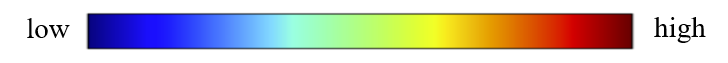}}\\
    \begin{tabular}{>{\centering\arraybackslash}m{0.4cm}>{\centering\arraybackslash}m{2.0cm}>{\centering\arraybackslash}m{2.0cm}>{\centering\arraybackslash}m{2.0cm}>{\centering\arraybackslash}m{2.0cm}}
        (a)&
        \includegraphics[width=2.0cm]{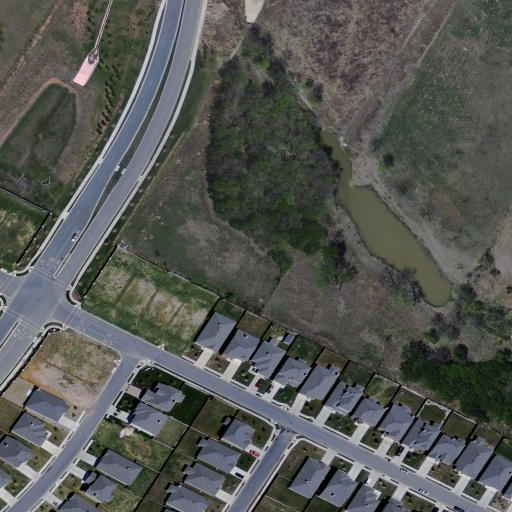} &
        \includegraphics[width=2.0cm]{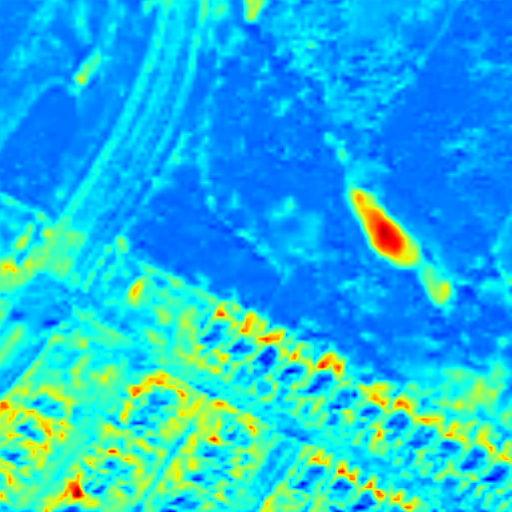} &
        \includegraphics[width=2.0cm]{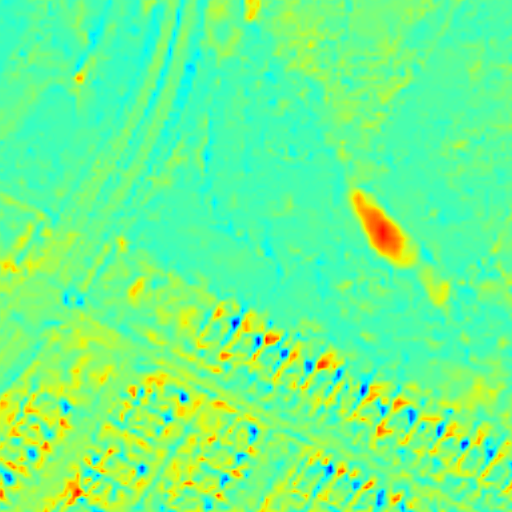} &
        \includegraphics[width=2.0cm]{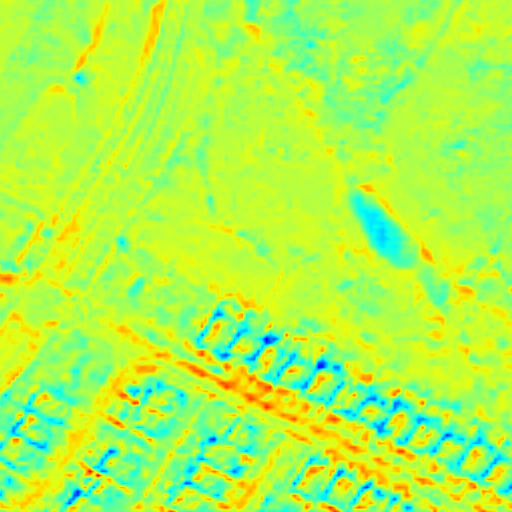} \\
        (b)&
        \includegraphics[width=2.0cm]{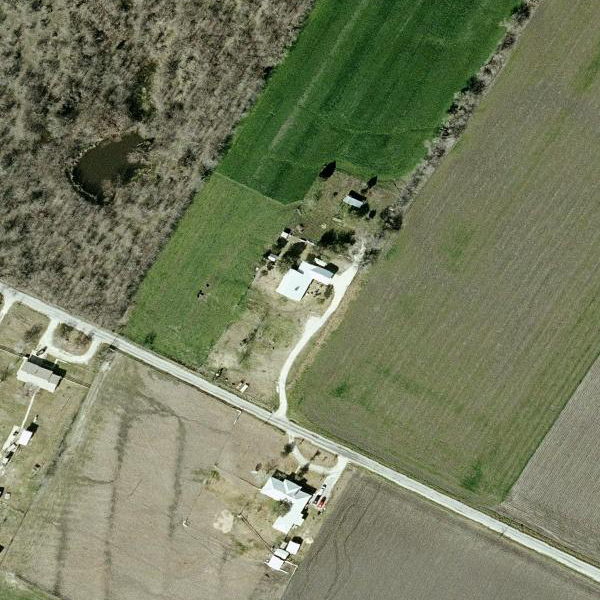} &
        \includegraphics[width=2.0cm]{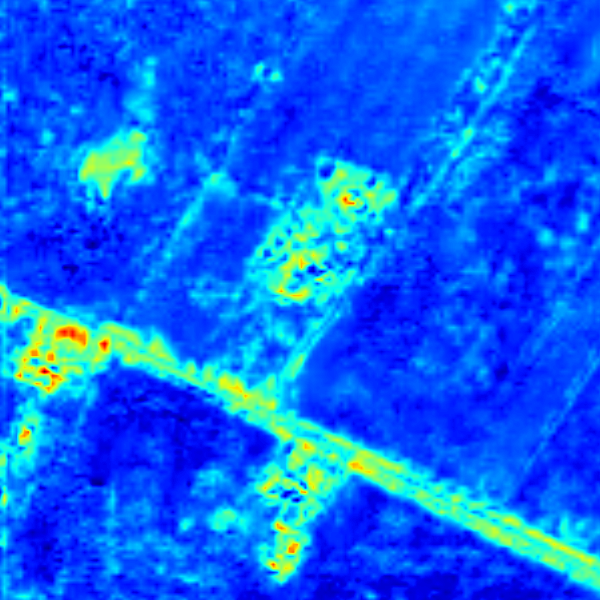} &
        \includegraphics[width=2.0cm]{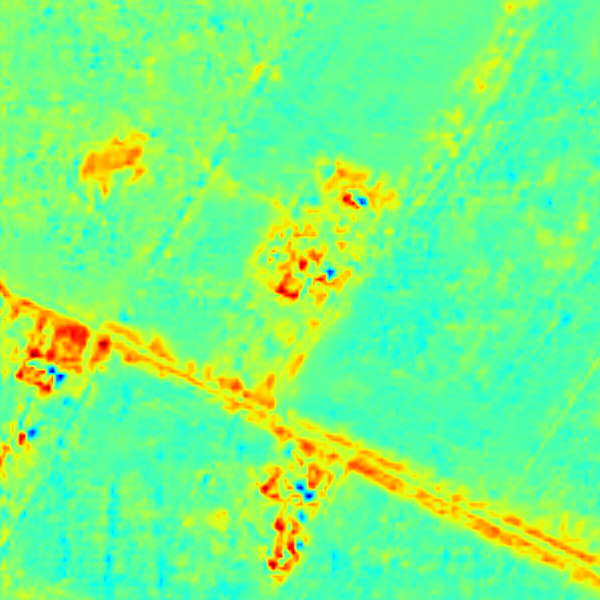} &
        \includegraphics[width=2.0cm]{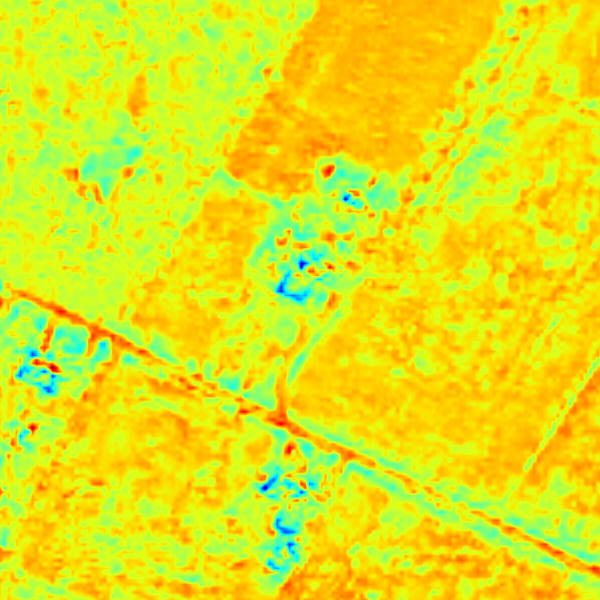} \\
        & \multirow{2}*{test RSI} & \multicolumn{3}{c}{$\underbrace{\: \: \: \: \: \: \: \: \: \: \: \: \: \: \: \: \: \: \: \: \: \: \: \: \: \: \: \: \: \: \: \: \: \: \: \: \: \: \: \: \: \: \: \: \: \: \: \: \: \: \: \: \: \: \: \: \: \: \: \: \: \: \: \: \: \: \: \: \: \: \: \:}$} \\
        & & \multicolumn{3}{c}{visualized semantic latent} \\
    \end{tabular}
    \caption{Visualization of the semantic latent. Warm colors indicate high values and vice versa for cold colors.}
    \label{Fig.latent}
\end{figure}

\textbf{Visualization of the latent:} To intuitively evaluate the effect of the proposed task-agnostic semantic learning, we visualize the semantic latent $l$ in equation (\ref{form.lt}) and present it in Fig.\ref{Fig.latent}. One can observe that the learned latent corresponds to various types of LCLU types that are not provided with semantic labels. Apart from buildings that are annotated in the CD datasets, semantic objects such as ponds, roads and low vegetation are attended to in the response maps. This demonstrates that the proposed method can capture the underlying semantics in RSIs, even without providing the corresponding human annotations.

\subsection{Comparative experiments with Fully-supervised CD Methods}\label{sc5.compareSOTA}

\textbf{Quantitative Results:} 
 We compare the accuracy of the proposed SAM-CD with the SOTA CD methods and report the results obtained in 4 benchmark datasets in Tables \ref{Table.CompareSOTA},\ref{Table.Compare_CLCD}, and \ref{Table.Compare_S2Look}. The compared methods include classic CNN-based methods such as FC-Siam-diff, FC-Siam-conc\cite{daudt2018fully} and SNUNet\cite{fang2021snunet}, and recent models based on transformer or self-attention such as BIT\cite{chen2021remote}, ChangeFormer\cite{bandara2022transformer}, CTD-Former\cite{zhang2023relation}, CGNet\cite{han2023change} and EATDer\cite{ma2023eatder}.

Among the literature methods, the recent transformer-based methods (including the BIT, ChangeFormer, CTD-Former, and EATDer) generally obtain higher accuracy. The CTDFormer obtains the 3rd-best mIoU on the WHU-CD and the CLCD dataset, while the EATDer and ChangeFormer obtain the 3rd-best accuracy on the Levir-CD and S2Looking datasets, respectively. CGNet outperforms most of the literature methods in all 4 datasets. It also obtains the highest \textit{Pre} and the highest \textit{Rec} on the WHU-CD and S2Looking datasets, respectively. However, a common problem of these literature methods is that they are mostly designed for detecting building changes, thus their capability of modeling LCLU changes and complex change scenes with spatial shift is limited. This can be verified in their relatively lower accuracy in the CLCD and S2Looking datasets.

The proposed SAM-CD achieves robust accuracy improvements over the SOTA methods in all 4 datasets. In particular, in the results obtained on the CLCD dataset, it surpass the SOTA accuracy by $1.31\%$ and $1.7\%$ in $mF_1$ and $mIoU$, respectively. This demonstrates that the semantic learning in SAM-CD, driven by the VFMs, helps to better discriminate the LCLU changes in RS scenes. In the WHU-CD and S2Looking datasets, it also outperforms the compared literature methods by a large margin. It obtains the best \textit{Rec} in the Levir-CD, WHU-CD and CLCD datasets, which indicates its capability to detect non-salient semantic changes.

\begin{table*}[t]
    \centering
    \caption{Quantitative results of different CD methods obtained on the Levir-CD and WHU-CD datasets (\%).}
    \resizebox{1\linewidth}{!}{%
        \begin{tabular}{r|ccccc|ccccc}
        \toprule
            \multirow{2}*{Method} & \multicolumn{5}{c}{Levir-CD} & \multicolumn{5}{c}{WHU-CD} \\
            \cline{2-11}
            & $Pre$ & $Rec$ & $OA$ & $mF_{1}$ & $mIoU$ & $Pre$ & $Rec$ & $OA$ & $mF_{1}$ & $mIoU$ \\
            \hline
            FC-Siam-diff\cite{daudt2018fully} & 92.93 & 87.02 & 98.15 & 89.73 & 82.67 & 94.52 & 89.32 & 98.70 & 91.74 & 85.65\\
            FC-Siam-conc\cite{daudt2018fully} & 92.03 & 89.82 & 98.28 & 90.89 & 84.34 & 93.94 & 92.83 & 98.90 & 93.37 & 88.19\\
            SNUNet\cite{fang2021snunet} & 93.83 & 90.11 & 98.50 & 91.88 & 85.84 & 92.61 & 83.26 & 98.10 & 87.30 & 79.40\\
            BIT\cite{chen2021remote} & 90.27 & 83.37 & 97.60 & 86.46 & 78.23 & 87.92 & 93.41 & 98.29 & 90.46 & 83.72\\
            ChangeFormer\cite{bandara2022transformer} & 91.46 & 86.31 & 97.95 & 88.69 & 81.21 & 96.28 & 92.95 & 99.12 & 94.55 & 90.08\\
            CTD-Former\cite{zhang2023relation} & 93.60 & 91.85 & 98.62 & 92.71 & 87.11 & 96.74 & 97.03 & 99.50 & 96.89 & 94.11\\
            CGNet\cite{han2023change} & 95.95 & 94.95 & 99.13 & 95.44 & 91.58 & \textbf{98.04} & 96.14 & 99.52 & 97.07 & 94.44\\
            EATDer\cite{ma2023eatder} & \textbf{96.29} & 92.70 & 98.87 & 94.41 & 89.84 & 95.57 & 93.06 & 99.01 & 94.28 & 89.64 \\
            \hline
            SAM-CD (proposed) & 95.87 & \textbf{95.14} & \textbf{99.14} & \textbf{95.50} & \textbf{91.68} & 97.97 & \textbf{97.20} & \textbf{99.60} & \textbf{97.58} & \textbf{95.36} \\
        \bottomrule
        \end{tabular}
        }\label{Table.CompareSOTA}
\end{table*}

\begin{table}[t]
    \centering
    \caption{Quantitative results of different CD methods obtained on the CLCD dataset (\%).}
    \resizebox{1\linewidth}{!}{%
        \begin{tabular}{r|ccccc}
        \toprule
            \multirow{2}*{Method} & \multicolumn{5}{c}{Accuracy Metrics}  \\
            \cline{2-6}
            & $Pre$ & $Rec$ & $OA$ & $mF_{1}$ & $mIoU$ \\
            \hline
            FC-Siam-diff\cite{daudt2018fully} & 83.13 & 74.60 & 94.73 & 78.10 & 68.19 \\
            FC-Siam-conc\cite{daudt2018fully} & 81.65 & 79.94 & 94.84 & 80.77 & 70.98  \\
            SNUNet\cite{fang2021snunet} & 85.92 & 82.62 & 95.84 & 84.19 & 75.11  \\
            BIT\cite{chen2021remote} & 81.42 & 70.27 & 94.17 & 74.39 & 64.51  \\
            ChangeFormer\cite{bandara2022transformer} & 84.41 & 81.36 & 95.47 & 82.80 & 73.40 \\
            CTD-Former\cite{zhang2023relation} & 87.29 & 83.17 & 96.12 & 85.08 & 76.24\\
            CGNet\cite{han2023change} & 85.87 & 85.30 & 96.00 & 85.58 & 76.83 \\
            EATDer\cite{ma2023eatder} & 87.16 & 77.18 & 93.84 & 81.16 & 71.18 \\
            \hline
            SAM-CD (proposed) & \textbf{88.25} & \textbf{85.65} & \textbf{96.26} & \textbf{86.89} & \textbf{78.53}\\
        \bottomrule
        \end{tabular}
        }\label{Table.Compare_CLCD}
\end{table}

\begin{table}[t]
    \centering
    \caption{Quantitative results of different CD methods obtained on the S2Looking dataset (\%).}
    \resizebox{1\linewidth}{!}{%
        \begin{tabular}{r|ccccc|ccccc}
        \toprule
            \multirow{2}*{Method} & \multicolumn{4}{c}{Accuracy Metrics} \\
            \cline{2-5}
            & $Pre$ & $Rec$ & $F_{1}$ & $IoU$ \\
            \hline
            FC-Siam-diff\cite{daudt2018fully} & 83.49 & 32.32 & 46.60 & 30.38 \\
            FC-Siam-conc\cite{daudt2018fully} & 68.27 & 18.52 & 13.54 & - \\
            SNUNet\cite{fang2021snunet} & 45.26 & 50.60 & 47.78 & 31.39 \\
            BIT\cite{chen2021remote} &  70.26 & 56.53 & 62.65 & 45.62 \\
            ChangeFormer\cite{bandara2022transformer} & \textbf{72.82} & 56.13 & 63.39 & - \\
            CGNet\cite{han2023change} & 70.18 & \textbf{59.38} & 64.33 & 47.41\\
            EATDer\cite{ma2023eatder} & 65.85 & 54.74 & 59.78 & 42.64\\
            \hline
            SAM-CD (proposed) & 72.80 & 58.92 & \textbf{65.13} & \textbf{48.29}\\
        \bottomrule
        \end{tabular}
        }\label{Table.Compare_S2Look}
\end{table}

\begin{figure*}[!htb]
\centering
    {\includegraphics[height=0.5cm]{Pics/BNsegColorBar.png}}\\
    \setlength{\tabcolsep}{1pt}
    \begin{tabular}{>{\centering\arraybackslash}m{0.35cm}>{\centering\arraybackslash}m{1.35cm}>{\centering\arraybackslash}m{1.35cm}>{\centering\arraybackslash}m{1.35cm}>{\centering\arraybackslash}m{1.35cm}>{\centering\arraybackslash}m{1.35cm}>{\centering\arraybackslash}m{1.35cm}>{\centering\arraybackslash}m{0.35cm}>{\centering\arraybackslash}m{1.35cm}>{\centering\arraybackslash}m{1.35cm}>{\centering\arraybackslash}m{1.35cm}>{\centering\arraybackslash}m{1.35cm}>{\centering\arraybackslash}m{1.35cm}>{\centering\arraybackslash}m{1.35cm}}
        (a)&
        \includegraphics[width=1.35cm]{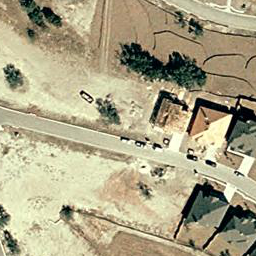} &
        \includegraphics[width=1.35cm]{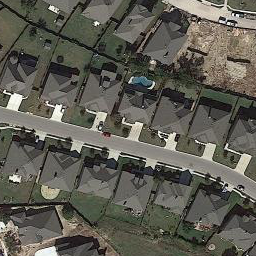} &
        \includegraphics[width=1.35cm]{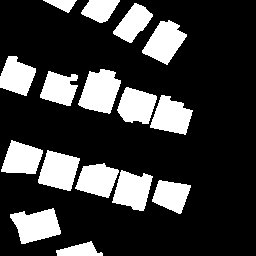} &
        \includegraphics[width=1.35cm]{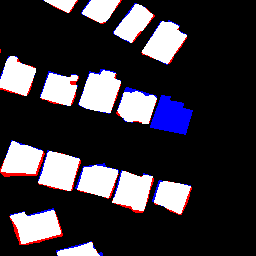} &
        \includegraphics[width=1.35cm]{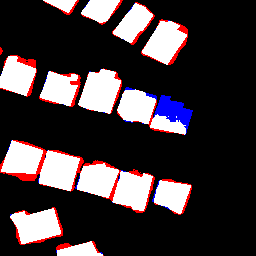} &
        \includegraphics[width=1.35cm]{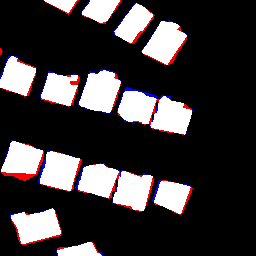} &
        (b)&
        \includegraphics[width=1.35cm]{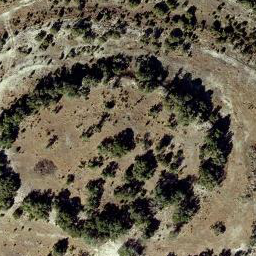} &
        \includegraphics[width=1.35cm]{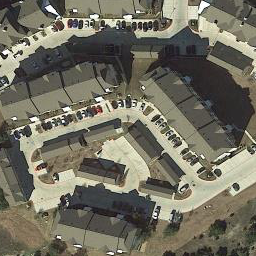} &
        \includegraphics[width=1.35cm]{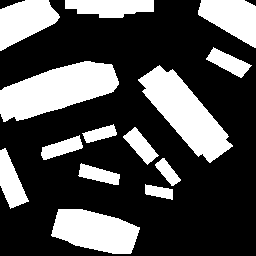} &
        \includegraphics[width=1.35cm]{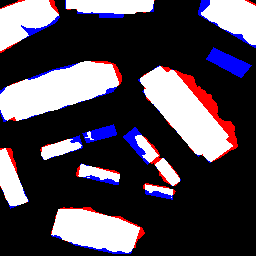} &
        \includegraphics[width=1.35cm]{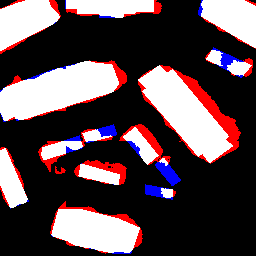} &
        \includegraphics[width=1.35cm]{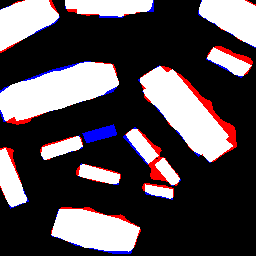}\\
        (c)&
        \includegraphics[width=1.35cm]{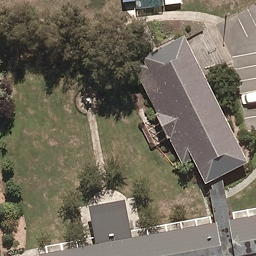} &
        \includegraphics[width=1.35cm]{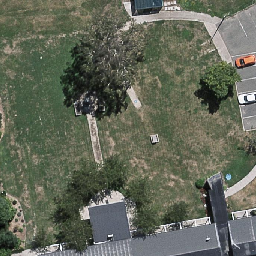} &
        \includegraphics[width=1.35cm]{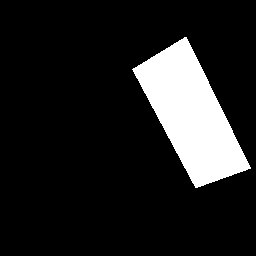} &
        \includegraphics[width=1.35cm]{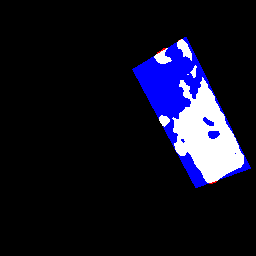} &
        \includegraphics[width=1.35cm]{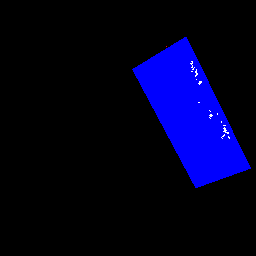} &
        \includegraphics[width=1.35cm]{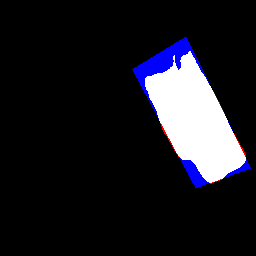} &
        (d)&
        \includegraphics[width=1.35cm]{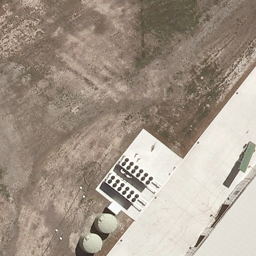} &
        \includegraphics[width=1.35cm]{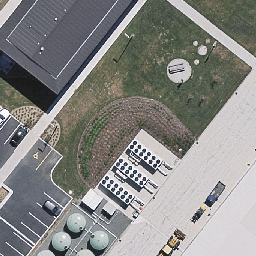} &
        \includegraphics[width=1.35cm]{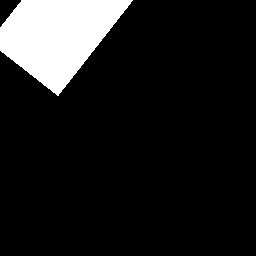} &
        \includegraphics[width=1.35cm]{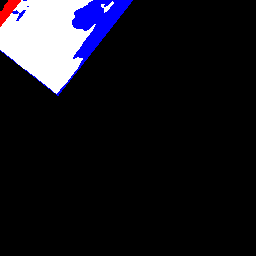} &
        \includegraphics[width=1.35cm]{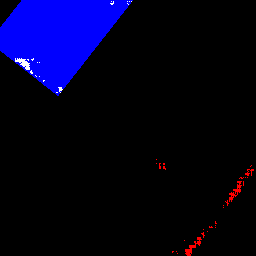} &
        \includegraphics[width=1.35cm]{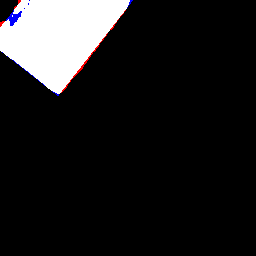}\\
        (e)&
        \includegraphics[width=1.35cm]{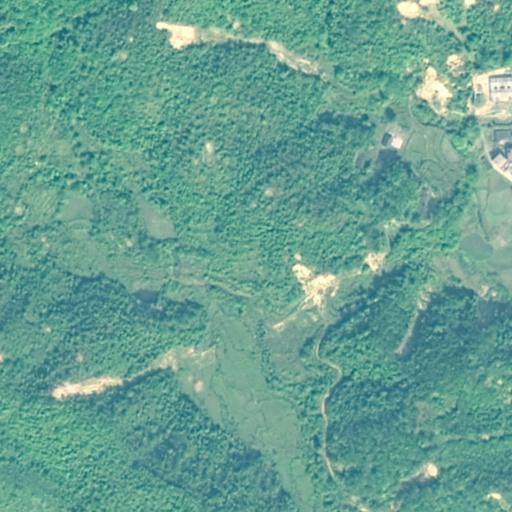} &
        \includegraphics[width=1.35cm]{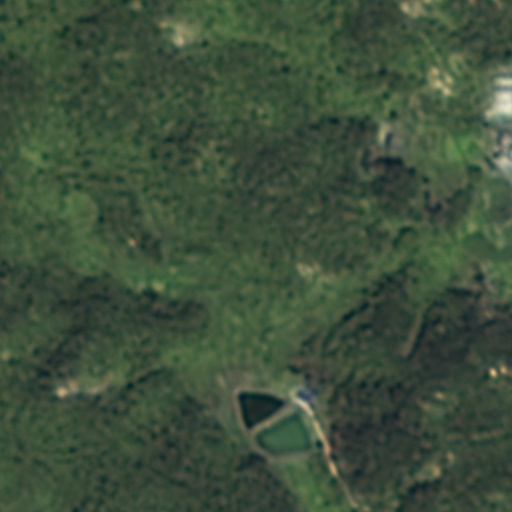} &
        \includegraphics[width=1.35cm]{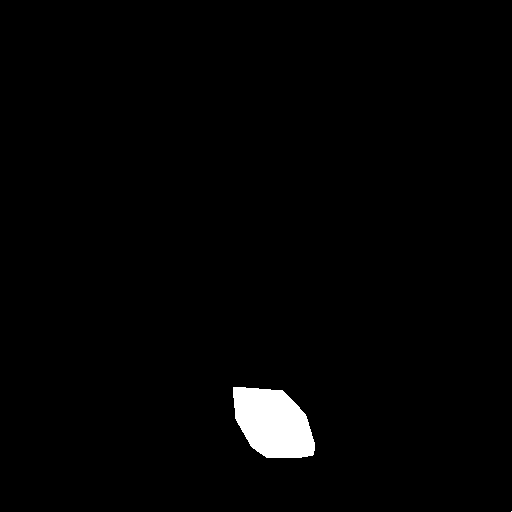} &
        \includegraphics[width=1.35cm]{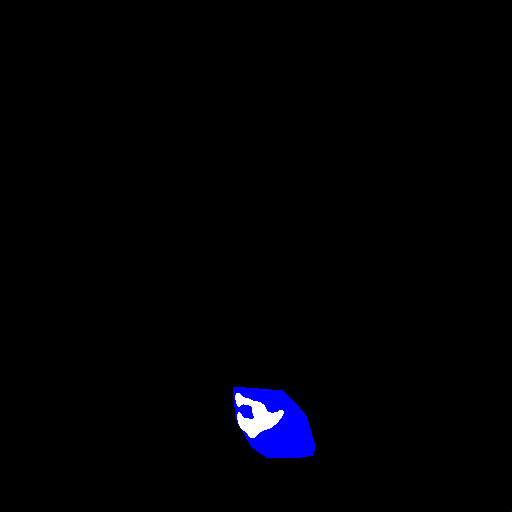} &
        \includegraphics[width=1.35cm]{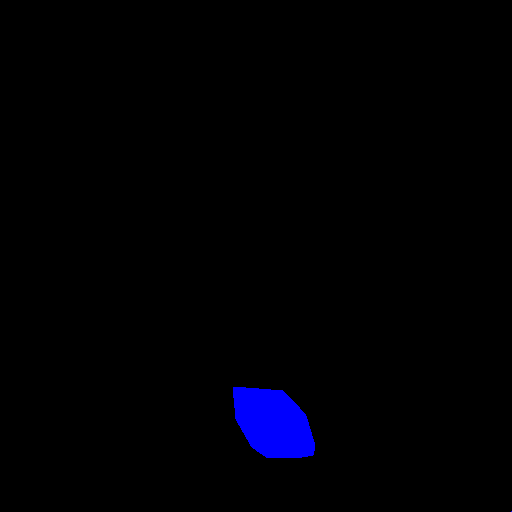} &
        \includegraphics[width=1.35cm]{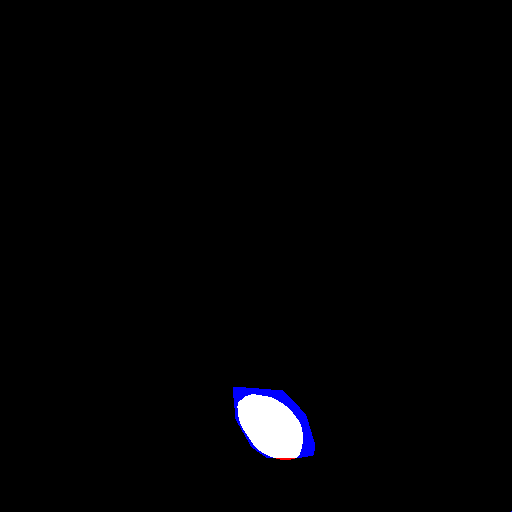} &
        (f)&
        \includegraphics[width=1.35cm]{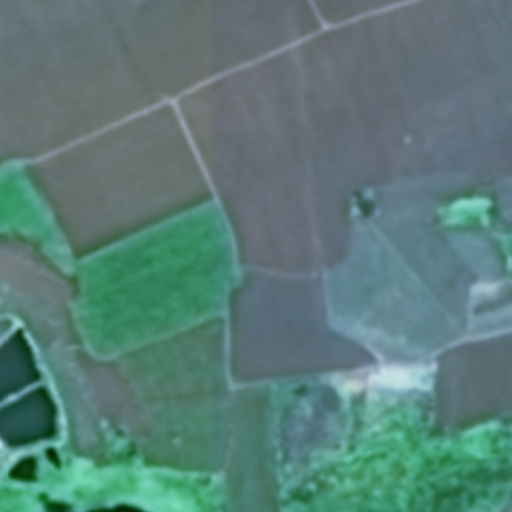} &
        \includegraphics[width=1.35cm]{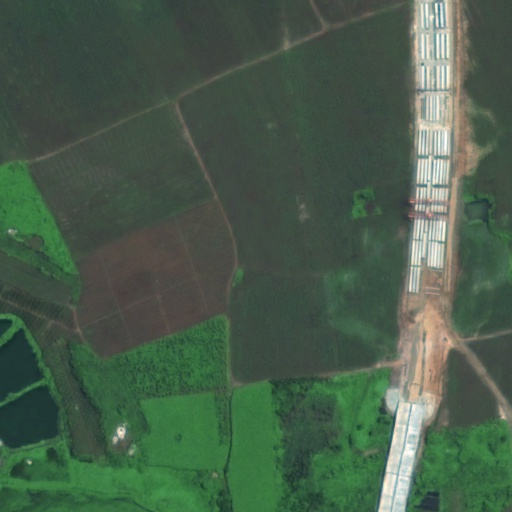} &
        \includegraphics[width=1.35cm]{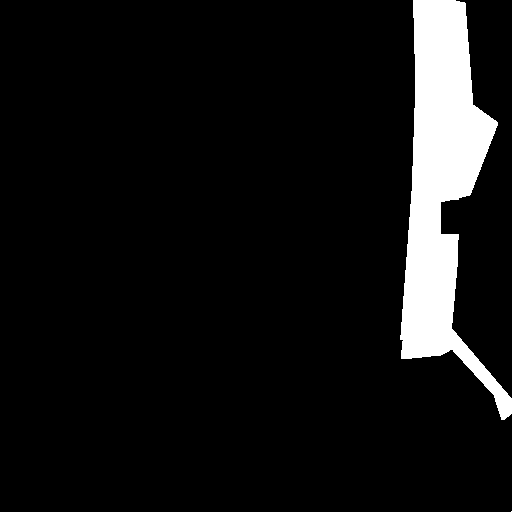} &
        \includegraphics[width=1.35cm]{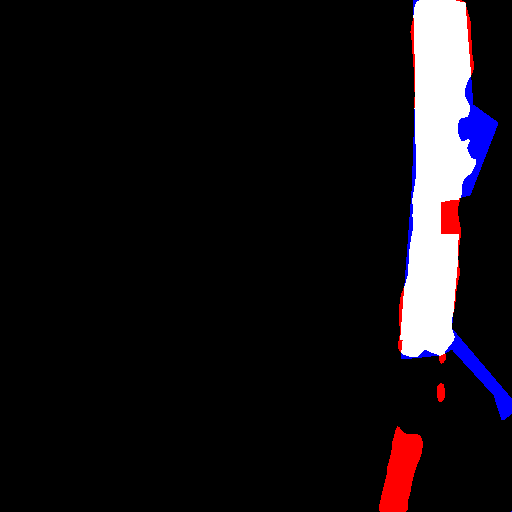} &
        \includegraphics[width=1.35cm]{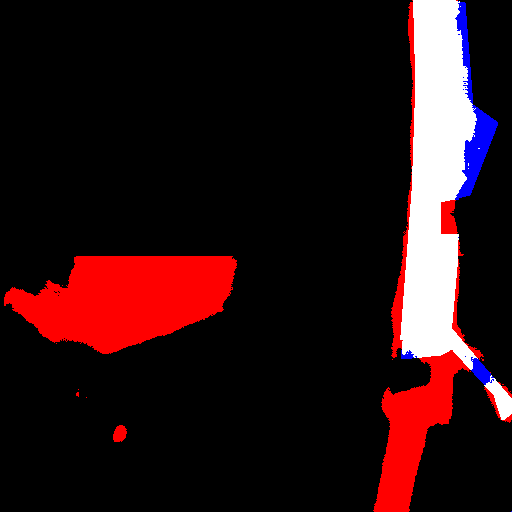} &
        \includegraphics[width=1.35cm]{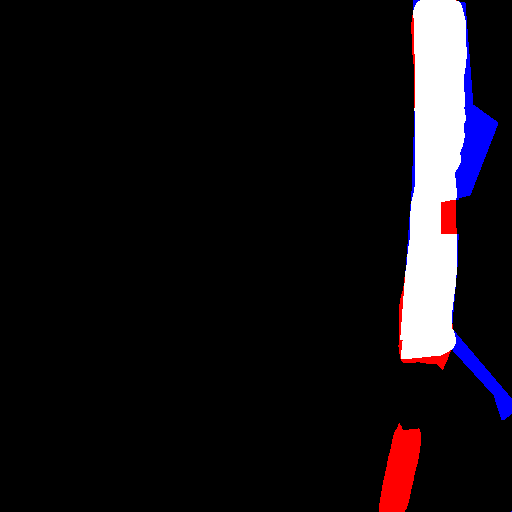}\\
        (g)&
        \includegraphics[width=1.35cm]{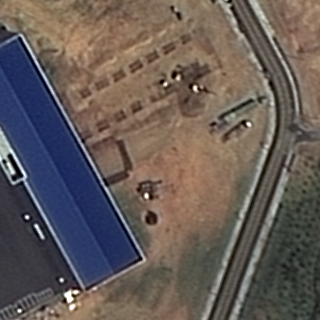} &
        \includegraphics[width=1.35cm]{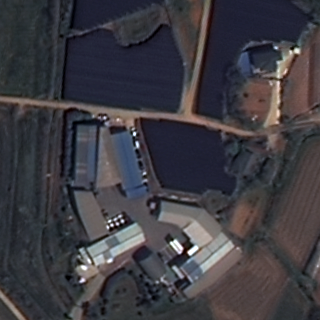} &
        \includegraphics[width=1.35cm]{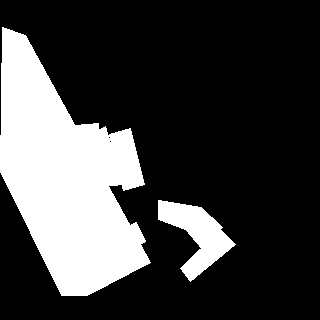} &
        \includegraphics[width=1.35cm]{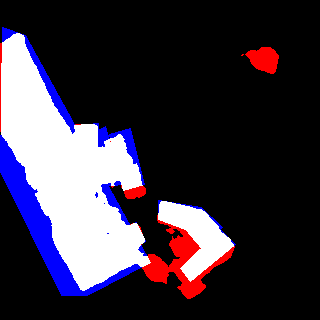} &
        \includegraphics[width=1.35cm]{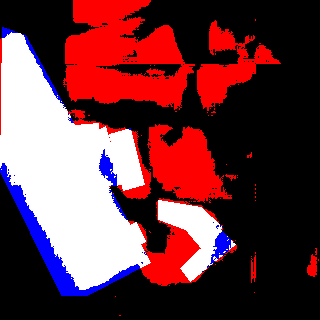} &
        \includegraphics[width=1.35cm]{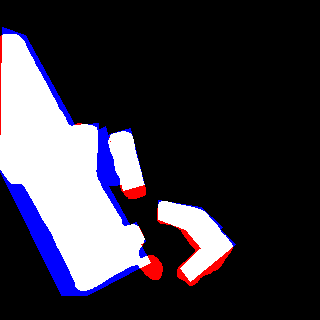} &
        (h)&
        \includegraphics[width=1.35cm]{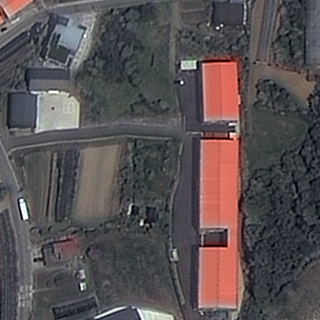} &
        \includegraphics[width=1.35cm]{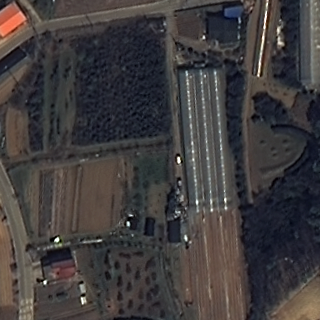} &
        \includegraphics[width=1.35cm]{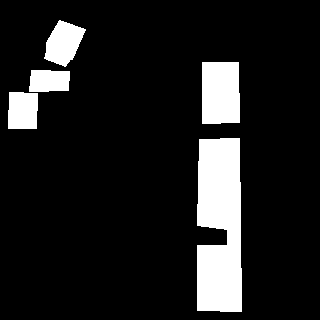} &
        \includegraphics[width=1.35cm]{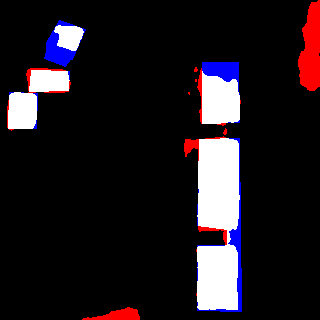} &
        \includegraphics[width=1.35cm]{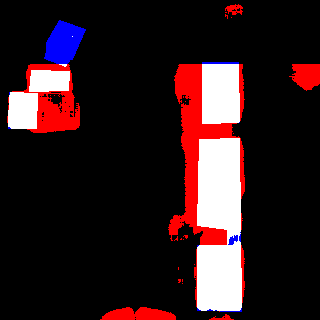} &
        \includegraphics[width=1.35cm]{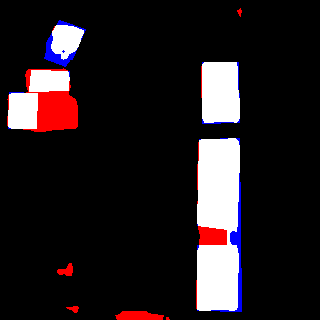}\\
        & $T_1$ image & $T_2$ image & GT & CGNet & EATDer & SAM-CD & & $T_1$ image & $T_2$ image & GT & CGNet & EATDer & SAM-CD\\
    \end{tabular}
    \caption{CD results of the different fully-supervised methods. (a)(b) results on the Levir-CD dataset, (c)(d) results on the WHU-CD dataset, (e)(f) results on the CLCD dataset, (g)(h) results on the S2Looking dataset.}
    \label{Fig.Compar_Results}
\end{figure*}

\textbf{Qualitative Results:} In Fig.\ref{Fig.Compar_Results} we present some examples of the CD results obtained on the 4 benchmark datasets. Compared with two recent methods CGNet and EATDer, the proposed SAM-CD exhibits advantages in detecting non-salient changes, such as new buildings (a)(d), removed house (c), new pond (e), and constructions (f). It also has fewer false alarms while dealing with complex changes in (b)(e)(h). This demonstrates that the SAM-CD can better discriminate between temporal differences and semantic changes.

\textbf{Computational Costs:} Table \ref{Table.CompCosts} reports the cost of computational resources of different methods. Among the recent literature methods, the SNUNet has the fewest parameters and the highest inference speed (indicated in infer.). The inference speed of the BIT and CGNet is also relatively high. In the SAM-CD, the FastSAM encoder accounts for most of the parameters (68m), whereas the proposed network modules account for 2.49m parameters. The computational efficiency (indicated in FLOPs) of the SAM-CD is much lower than most of the recent methods (including SNUNet, BIT, ChangeFormer, CTD-Former, CGNet and EATDer). However, the inference speed of the SAM-CD is higher than most of the recent methods except for the ChangeFormer.

\begin{table}[t]
    \centering
    \caption{The computational costs of different CD methods.}
    \resizebox{1\linewidth}{!}{%
        \begin{tabular}{r|ccc}
        \toprule
            Method & Param(Mb) & FLOPs(Gps) & Infer.(s)\\
            \hline
            FC-Siam-diff & 1.66 & 21.45 & 0.66 \\ 
            FC-Siam-conc & 2.74 & 8.77 & 0.56 \\
            SNUNet & 1.35 & 18.91 & 0.79 \\
            BIT & 11.47 & 19.60 & 1.12 \\
            ChangeFormer & 41.03 & 811.15 & 10.77 \\
            CTD-Former & 3.85 & 104.62 & 2.62 \\
            CGNet & 33.68 & 82.23 & 1.54\\ 
            EATDer & 6.61 & 23.46 & 3.67\\
            \hline
            \multirow{2}*{SAM-CD} & 2.49 & \multirow{2}*{8.60} & \multirow{2}*{4.59}\\
            & +68 (FastSAM)\\
        \bottomrule
        \end{tabular}
        }\label{Table.CompCosts}
\end{table}

\subsection{Comparative experiments with Semi-supervised CD Methods}

A valuable feature of CFMs is their generalization to visual tasks. In experiments, we find that the proposed SAM-CD exhibits less dependence on training samples compared to standard deep learning-based methods. Table \ref{Table.Compare_semiCD_Levir} and \ref{Table.Compare_semiCD_WHU} compare the accuracy of different CD methods trained with limited proportions of training data. Apart from the baseline method (ResNet-CD), we also compare with the SOTA methods for semi-supervised CD, including S4GAN\cite{mittal2019semi}, SemiCDNet\cite{peng2020semicdnet}, SemiCD\cite{bandara2022revisiting} and UniMatch\cite{yang2023revisiting}. It is worth noting that the reported accuracy of the SAM-CD is obtained using plain supervised training, i.e., without using complex semi-supervised training strategies such as contrastive learning, adversarial learning, and spectral-wise augmentations. The reported accuracy in this section is obtained following the experimental setups in \cite{bandara2022revisiting}.

\begin{table*}[t]
    \centering
    \caption{Accuracy of the SAM-CD and semi-supervised CD methods versus different proportions of training data (tested on Levir-CD). The evaluation metrics are $IoU^c$ and OA, respectively.}
    \resizebox{0.9\linewidth}{!}{%
        \begin{tabular}{r|ccccc}
        \toprule
            \multirow{2}*{Methods} & \multicolumn{5}{c}{Levir-CD}\\
            \cline{2-6}
            & 5\% & 10\% & 20\% & 40\% & 100\%\\
            \hline
            S4GAN \cite{mittal2019semi}& 64.0 / 97.89 & 67.0 / 98.11 & 73.4 / 98.51 & 75.4 / 98.62 & - \\
            SemiCDNet \cite{peng2020semicdnet} & 67.6 / 98.17 & 71.5 / 98.42 & 74.3 / 98.58 & 75.5 / 98.63 & -\\
            SemiCD \cite{bandara2022revisiting} & 72.5 / 98.47 & 75.5 / 98.63 & 76.2 / 98.68 & 77.2 / 98.72 & 77.9 / 98.77 \\
            UniMatch \cite{yang2023revisiting}& \textbf{80.7 / 98.95} & \textbf{82.0 / 99.02} & \textbf{81.7 / 99.02} & \underline{82.1 / 99.03} & - \\
            \hline
            SAM-CD (proposed) & \underline{73.87 / 98.50} & \underline{79.17 / 98.86} & \underline{80.18 / 98.91} & \textbf{83.17} / \textbf{99.07} & \textbf{84.26} / \textbf{99.14} \\
        \bottomrule
        \end{tabular} }\label{Table.Compare_semiCD_Levir}
\end{table*}

\begin{table*}[t]
    \centering
    \caption{Accuracy of the SAM-CD and semi-supervised CD methods versus different proportions of training data (tested on WHU-CD). The evaluation metrics are $IoU^c$ and OA, respectively.}
    \resizebox{0.9\linewidth}{!}{%
        \begin{tabular}{r|ccccc}
        \toprule
            \multirow{2}*{Methods} & \multicolumn{5}{c}{WHU-CD} \\
            \cline{2-6}
            & 5\% & 10\% & 20\% & 40\% & 100\% \\
            \hline
            S4GAN \cite{mittal2019semi}& 18.3 / 96.69 & 62.6 / 98.15 & 70.8 / 98.60 & 76.4 / 98.96 & -\\
            SemiCDNet \cite{peng2020semicdnet} & 51.7 / 97.71 & 62.0 / 98.16 & 66.7 / 98.28 & 75.9 / 98.93 & -\\
            SemiCD \cite{bandara2022revisiting} & 65.8 / 98.37 & 68.1 / 98.47 & 74.8 / 98.84 & 77.2 / 98.96 & 85.5 / 99.38\\
            UniMatch \cite{yang2023revisiting} & \textbf{80.2 / 99.15} & \textbf{81.7 / 99.22} & \textbf{81.7 / 99.18} & \textbf{85.1 / 99.35} & - \\
            \hline
            SAM-CD (proposed) & \underline{ 66.91 / 98.56 } & \underline{71.38 / 98.73} & \underline{77.92 / 99.00} & \underline{84.00 / 99.23} & \textbf{91.15} / \textbf{99.60}\\
        \bottomrule
        \end{tabular} }\label{Table.Compare_semiCD_WHU}
\end{table*}

\begin{figure}[t]
\centering
    \includegraphics[width=1\linewidth]{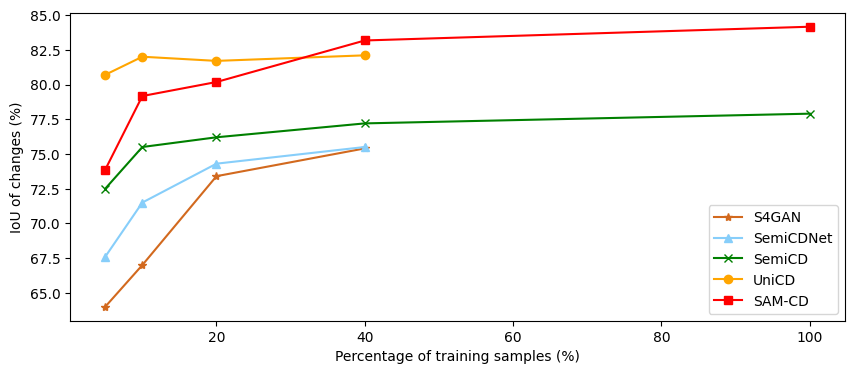}
    \caption{The $IoU_c$ of semi-supervised CD methods versus different percentages of training data (tested on the Levir-CD dataset).}
    \label{Fig.semi_acc_levir}
\end{figure}

\begin{figure}[t]
\centering
    \includegraphics[width=1\linewidth]{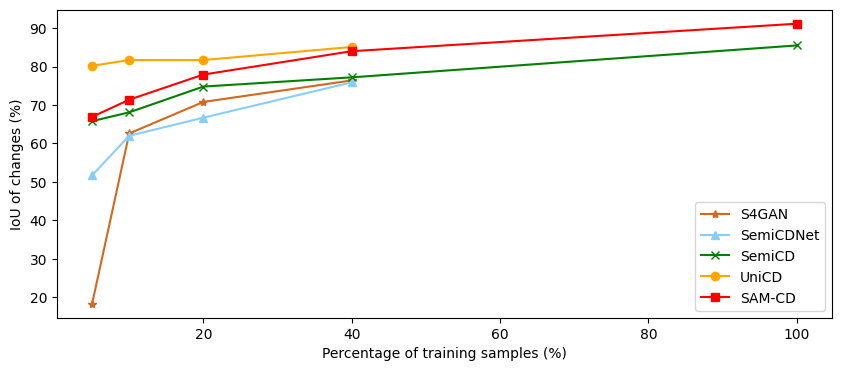}
    \caption{The $IoU_c$ of semi-supervised CD methods versus different percentages of training data (tested on the WHU-CD dataset).}
    \label{Fig.semi_acc_whu}
\end{figure}

The quantitative results (reported in Table \ref{Table.Compare_semiCD_Levir} and Table \ref{Table.Compare_semiCD_WHU}) reveal that the SAM-CD outperforms most semi-supervised CD methods, except for the very recent method UniMatch \cite{yang2023revisiting}. Due to its contrastive learning and feature-level augmentation strategies, the UniMatch obtains the highest accuracy with 5\%, 10\% and 20\% of the training data. The proposed SAM-CD obtains the highest accuracy while using 40\% and 100\% of the training data (on the Levir-CD dataset), and obtains the second-best accuracy with 5\%, 10\% and 20\% of the training data. However, there is a drop in accuracy while using only 5\% of the training samples. We assume that this is because the training of the adaptor and the change head still requires a certain extent of supervised training. We further plot the accuracy versus different training proportions in Fig.\ref{Fig.semi_acc_levir} and Fig.\ref{Fig.semi_acc_whu}. The evaluation metric is the $IoU$ of change areas following the literature studies \cite{yang2023revisiting}. It points out that, while a certain number of training samples are still required to adapt the FastSAM to CD task, the accuracy of the SAM-CD is comparable to SOTA methods for semi-supervised CD.
\section{Conclusion}\label{sc5}

Typical deep learning-based CD methods compare temporal differences to segment the interesting changes, thus they are often affected by seasonal changes and different imaging conditions. In this paper, we propose a SAM-CD architecture that models the semantic latent in VHR RSIs to detect the changed objects. It leverages the FastSAM model to extract visual features in ground objects, and utilizes the underlying temporal constraints in RSIs to supervise the learning of task-agnostic semantic representations.

Experimental results tested on 4 benchmark datasets demonstrate that the proposed SAM-CD gains significant accuracy improvements over the SOTA methods. Due to its capability of modeling semantic latent in VHR RSIs, it can better discriminate between semantic changes and temporal differences. Moreover, benefiting from the great generalization capability of VFM encoders, the SAM-CD exhibits a sample-efficient learning ability. It can obtain fairly accurate results with a limited number of training samples.

However, there are still several remaining problems. First, the proposed SAM-CD still requires fully-supervised training. Under the circumstances that there are only very limited training samples, its accuracy is below the SOTA methods in semi-supervised CD. Second, the inference speed of the SAM-CD is relatively slow compared with literature methods. To better utilize VFMs in CD tasks, there is a need to reduce the parameters and discard the redundant priors. This can be achieved by using knowledge-distilling learning techniques.

With the advances in VFMs, it is now possible to reach zero-shot or few-shot CD in VHR RSIs with high accuracy. In the future, we will continue to explore combining VFM-based CD with self-supervised training techniques to approach this goal.

\bibliographystyle{IEEEtran}
\bibliography{refs}

\end{document}